\documentclass[letterpaper,12pt]{article}
\usepackage{hyperref} 
\usepackage{osajnl2}

\usepackage{epsfig,amsmath,amssymb,amssymb,graphicx,subfig,color,multirow}
\usepackage[titletoc]{appendix}

\newtheorem{theorem}{Theorem}

\graphicspath{{figures/}}

\begin{document}

\title{High-precision camera distortion measurements
       with a ``calibration harp''}
\author{Z.~Tang,$^{1,*}$ R.~Grompone von Gioi,$^2$
        P.~Monasse,$^3$ and J.M.~Morel$^2$}
\address{$^1$ECE, University of Minnesota, Minneapolis, USA}
\address{$^2$CMLA, ENS Cachan, France}
\address{$^3$IMAGINE, LIGM, Univ. Paris-Est/\'Ecole des Ponts ParisTech, France}
\address{$^*$Corresponding author: tang@cmla.ens-cachan.fr}

\begin{abstract}
This paper addresses the high precision measurement of the distortion
of a digital camera from photographs. Traditionally, this distortion
is measured from photographs of a flat pattern which contains aligned
elements. Nevertheless, it is nearly impossible to fabricate a very
flat pattern and to validate its flatness. This fact limits the
attainable measurable precisions. In contrast, it is much easier to
obtain physically very precise straight lines by tightly stretching
good quality strings on a frame. Taking literally ``plumb-line
methods'', we built a ``calibration harp'' instead of the classic flat
patterns to obtain a high precision measurement tool, demonstrably
reaching $2/100$ pixel precisions. The harp is complemented with the
algorithms computing automatically from harp photographs two different
and complementary lens distortion measurements. The precision of the
method is evaluated on images corrected by state-of-the-art distortion
correction algorithms, and by popular software. Three applications are
shown: first an objective and reliable measurement of the result of
any distortion correction. Second, the harp permits to control
state-of-the art global camera calibration algorithms: It permits to
select the right distortion model, thus avoiding internal compensation
errors inherent to these methods. Third, the method replaces manual
procedures in other distortion correction methods, makes them fully
automatic, and increases their reliability and precision.

\end{abstract}

\begin{quote}
Note to referees and editor: Our lens distortion measurement algorithm
can be tested on the online demo version available at
\url{http://bit.ly/lens-distortion}.
\end{quote}

\section{Introduction}\label{sec:introduction}

The precision of 3D stereovision applications is intimately related to
the precision of the camera calibration, and especially of the camera
distortion correction. An imprecise distortion model produces residual
distortion that will be directly back-projected to the reconstructed
3D scene. Such imprecision can be a serious hindrance in remote
sensing applications such as the early warnings of geology disasters,
or in the construction of topographic maps from stereographic pairs of
aerial photographs. The fast growing resolution of digital cameras and
of their optical quality is transforming them into (potential) high
precision measurement tools. Thus, it becomes important to measure the
calibration precision with ever higher precision. 

A first step toward high-precision distortion corrections is to
perform precise distortion measurements. This basic tool can then be
used to evaluate the precision of a correction method, or can become
part of the correction method itself.

\begin{figure}[!htb]
\centering
\includegraphics[width=.4\textwidth]{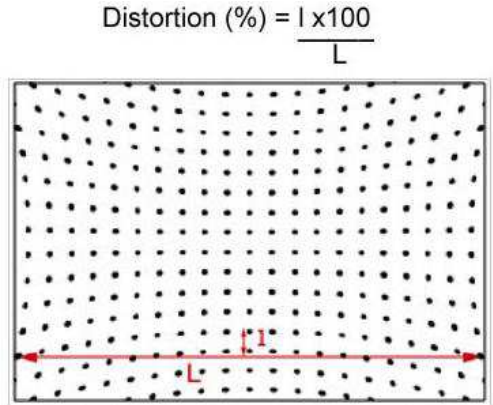}
\caption{DxO lens distortion measurement standard. Estimation of
  distortion from an image of a dot chart.}
\label{fig:dxo_distortion}
\end{figure}

Camera and lens distortion measurement methods usually require a flat
pattern containing aligned elements. The pattern is photographed using
the target lens, and the distortion is measured by how much the
observed elements deviate from the straight alignment on the
pattern. For example the classic DxO-labs' software, a good
representative of camera maker practice, (\url{http://www.dxo.com/})
uses a pattern with a grid of aligned dots. Distortion is measured by
the positional errors associated with the maximal deviation in a row,
see Fig.~\ref{fig:dxo_distortion}. Similar methods are proposed by the
SMIA\footnote{Standard Mobile Imaging Architecture standard},
EBU\footnote{European Broadcasting Union standard
  \url{http://www.ebu.ch/}}, IE\footnote{Image Engineering standard
  \url{http://www.image-engineering.de/}}, and
I3A\footnote{International Imaging Industry Association standard
  \url{http://www.i3a.org/}} standards. These measurements are
generally manual and require a perfectly flat pattern.

Every lens distortion correction method includes, implicitly, a lens
distortion measurement. These methods can be roughly classified into
four groups:
\begin{itemize}
  \item re-projection error minimization methods;
  \item pattern matching methods;
  \item enlarged epipolar geometry based methods;
  \item plumb-line methods.
\end{itemize}
Re-projection methods usually rely on a planar pattern containing
simple geometric shapes. In these methods, the lens distortion is
estimated together with the camera internal and external parameters
\cite{Slama1980, Tsai1987, Zhang2000, Lavest1998, Weng1992}, by
minimizing the re-projection error between the observed control points
on the pattern and the re-projected control points simulated by the
pattern model and camera model. The distortion is measured in terms of
re-projection error once all the parameters have been estimated. Both
the internal parameter errors and the external parameter errors
contribute to the re-projection error. Unfortunately, these errors can
compensate each other. Thus, a small re-projection error may be
observed while the internal and external parameters are not well
estimated. This compensation effect reduces the precision of the
estimation of the lens distortion parameters \cite{TheseZhongwei}. It
hinders accurate distortion estimation from the re-projection error. 
Notice that this high precision 
is not always required. In some applications, the distortion estimated in these 
bundle adjustment based re-projection methods, which makes the 3D geometry consistent 
with the pin-hole model, is precise enough for human observer.

A quite different kind of method proceeds by matching a photograph of
a flat pattern to its digital model. These methods estimate the
distortion field by interpolating a continuous distortion field from a
set of matching points. Several variants exist depending on the kind
of pattern, matching and interpolation technique. A common and crucial
assumption for these methods is that the pattern is flat. In practice,
however, it is difficult to produce a very flat pattern, and the
consequences of a tiny flatness flaw are considerable. For example it
is reported in \cite{Grompone2010} that a flatness error of about
100~$\mu$m for a 40 centimeters broad pattern can lead to an error of
about $0.3$ pixels in the distortion field computation for a Canon EOS
30D camera of focal length $18$~mm with the distance between the
camera and the object about $30$~cm. The only physical method to
assess a pattern flatness at a high precision is interferometry, but
it requires the pattern to be a mirror, which is not adequate for
photography. Furthermore, camera calibration requires large patterns,
which are therefore flexible. Deformations of the order of 100~$\mu$m
or more can be caused by temperature changes, and by a mere position
change of the pattern, which deforms under its own weight.

Recently, more attention has been paid to pattern-free methods (or
self-calibration methods) where the distortion estimation is obtained
without using any specific pattern. The distortion is estimated from
the correspondences between two or several images in absence of any
camera information. The main tool is the so-called enlarged epipolar
constraint, which incorporates lens distortion into the epipolar
geometry. Some iterative \cite{Stein1997, Zhang1996} or non-iterative
methods \cite{Fitzgibbon2001, Micusik2003, Li2005, Thirthala2005,
  Claus2005a, Barreto2005, Kukelova2007, Kukelova2008, Byrod2008,
  Kukelova2007a, Josephson2009} are used to estimate the distortion
and correct it by minimizing an algebraic error. The estimated
distortion can be used as the initialization in bundle adjustment
methods to improve the camera calibration precision \cite{Triggs2000}.

The so called ``plumb-line'' methods, which correct the distortion by
rectifying distorted images of 3D straight lines, date back to the
1970s (see Brown's seminal paper in 1971 \cite{Brown1971}). Since
then, this idea has been applied to many distortion models: the radial
model \cite{Alvarez2009, PRESCOTT1997, Pajdla1997}, the FOV (Field Of
View) model \cite{Devernay2001}, or the rational function model
\cite{Claus2005}. These methods minimize the straightness error of the
corrected lines. According to the fundamental theorem to be introduced
in section~\ref{sec:theorem}, the plumb-line methods minimize an error
directly related to the distortion, without suffering from the above
mentioned drawback, namely a numerical error compensation. On the
other hand, in spite of their name, current digital plumb-line methods
usually involve flat patterns with alignments on them and not the
plumb lines that were originally proposed in analog photogrammetry.

Taken literally, these methods should use photographs of 3D straight
lines. When a high precision is required, this setup becomes much
easier to build than a flat pattern. The main purpose of this article
is to show that very accurate distortion evaluation and correction can
be obtained with a basic plumb line tool which we called ``calibration
harp''. The calibration harp is nothing but a frame supporting tightly
stretched strings. Nevertheless, the photographs of a calibration harp
require a new numerical treatment to exploit them. The strings will
have to be detected at high sub-pixel accuracy and their distortion
converted into an invariant measurement.

Distortion measurements can be used to evaluate the distortion of a
camera, but also its residual distortion after correction. Two aspects
of the measurement should be clarified here. In this paper we discuss
distortion measurements that apply to the camera conceived as a whole:
It is impossible to tell which part the relative position and
deformation of the CCD, and the lens distortion itself play in the
global camera distortion. The distortion measurement is therefore not
a pure optical lens distortion measurement, but the distortion
measurement of the full acquisition system of camera+lens in a given
state. Different lenses on different cameras can be compared only when
the camera calibration matrix is known. On the other hand, the
residual error due to different correction algorithms can be compared
objectively after applying an appropriate normalization on the
corrected images.

This paper is organized as follows: The fundamental theorem
characterizing undistorted cameras is introduced in
section~\ref{sec:theorem}. Section~\ref{sec:harp_building} uncovers
the simple fabrication secrets of calibration harps. The image
processing algorithms needed for an automatic measurement are
presented in section~\ref{sec:edge_detection} and
section~\ref{sec:measurements} introduces the two most relevant
measures. Section~\ref{sec:applications} demonstrates two
applications, to the measurement of residual distortion after applying
a calibration method, and to the transformation of existing manual
distortion correction methods into automatic and far more precise
ones. Finally, section~\ref{sec:conclusion} concludes the paper.

\section{From Straight Lines to Straight Lines}\label{sec:theorem}

In multiple-view geometry, the pinhole camera is the ideal model that
all techniques tend to approximate at best by calibrating the real
cameras. This model corresponds to the ideal geometric perspective
projection. The next theorem characterizes perspective projections by
the fact that they preserve alignments. Instead of restating the
simplified version in \cite{Hartley2004}, we prefer to make it more
precise. The proof of the theorem can be found in
\cite{TheseZhongwei}.

\begin{theorem}\label{theo:fundamental_prop}
\it Let $\mathbf{T}$ be a continuous map from $\mathcal{P}^3$ to
$\mathcal{P}^2$ (from 3D projective space to 2D projective plane). If
there is a point $\mathbf{C}$ such that:
\begin{enumerate}
\renewcommand{\labelenumi}{(\alph{enumi})}

  \item the images of any three point belonging to a line in
    $\mathcal{P}^3$ not containing $\mathbf{C}$, are aligned points in
    $\mathcal{P}^2$;

  \item the images of any two points belonging to a line in
    $\mathcal{P}^3$ containing $\mathbf{C}$, are the same point in
    $\mathcal{P}^2$;

  \item there are at least four points belonging to a plane not
    containing $\mathbf{C}$, such that any group of three of them are
    non aligned, and their images are non aligned either;

\end{enumerate}
then $\mathbf{T}$ is a pinhole camera with center $\mathbf{C}$.
\end{theorem}

This theorem provides us with a fundamental tool to verify that a
camera follows the pinhole model. Nevertheless, \emph{rectifying
  straight lines does not define a unique distortion correction}: two
corrections can differ by any 2D homography that preserves all
alignments. More concretely, assume that the real camera model is
$\mathbf{P} = \mathcal{D}\mathbf{KR}[\mathbf{I}~ |~ -\mathbf{C}]$ with
$\mathbf{C}$ the coordinate of camera optic center in a given 3D world
frame, $\mathbf{R}$ the camera $3 \times 3$ orientation matrix,
$\mathbf{K}$ the camera $3 \times 3$ calibration matrix and
$\mathcal{D}$ the camera lens non-linear distortion. The estimated
distortion can be written as $\mathcal{D} \mathbf{H}^{-1}$ with
$\mathbf{H}^{-1}$ the unknown homography introduced in the distortion
correction and can be different from one correction algorithm to
another. By applying the inverse of the estimated distortion, the
recovered pinhole camera is $\hat{\mathbf{P}} =
\mathbf{H}\mathbf{KR}[\mathbf{I}~ |~ -\mathbf{C}]$. The homography
$\mathbf{H}$ can enlarge or reduce the straightness error, which makes
the comparison of different correction algorithms unfair. This effect
can be compensated by two strategies.

To arrive at a universal measurement, a first strategy that we will
consider is to normalize the homography:
\begin{enumerate}

  \item Select four points $\mathbf{P}_{i=1,\ldots,4}$ in the
    distorted image in general position (not three of them aligned).
    For example, they can be the four corners of the distorted image.

  \item Find their corresponding points $\mathbf{P}_{i}'$ in the
    corrected image, according to the correction model:
    $\mathbf{P}_{i}' = \mathbf{H}\mathcal{D}^{-1}
    \mathbf{P}_{i}$. Note that $\mathbf{H}$ is different from one
    correction algorithm to another.

  \item Compute the normalization homography $\hat{\mathbf{H}}$ which
    maps $\mathbf{P}_{i}'$ to $\mathbf{P}_{i}$: $\mathbf{P}_{i} =
    \hat{\mathbf{H}} \mathbf{P}_{i}'$. Note that $\hat{\mathbf{H}}$ is
    different from one correction algorithm to another.

  \item Apply the normalization homography $\hat{\mathbf{H}}$ on the
    corrected image.

\end{enumerate}
With this normalization, the final correction model is
$\hat{\mathbf{H}}\mathbf{H}\mathcal{D}^{-1}$.

A second possible strategy would be to fix specific parameters in the
correction model. For example, since the zoom factor in the distortion
correction is mainly determined by the order-$1$ parameters in the
correction model, it is sufficient to set all the order-$1$ parameters
to be $1$ to obtain a unique distortion measurement. Unfortunately,
this will not be possible for some non-parametric distortion
correction methods. The first strategy therefore is more general.

\section{Building a Calibration Harp}\label{sec:harp_building}

Theorem \ref{theo:fundamental_prop} suggests to take a set of physical
straight lines, as a calibration pattern. However, a common practice
actually contradicts the plumb-line basic idea: line patterns are
printed and pasted over a flat plate. There are many sources of
imprecision in this setup: the printer quality is not perfect; the
paper thickness is not perfectly uniform; the pasting process can add
bubbles or a non uniform glue layer; the supporting surface is not
perfectly flat either. Notwithstanding these defects, if only a pixel
precision is required, this setup is quite sufficient. Nonetheless,
when high sub-pixel precision is involved, the flatness errors cannot
be neglected. For current camera precision, a flatness error of
100~$\mu$m (the thickness of current writing paper) for a $40$~cm
pattern can lead to errors in the observed image coordinates of about
$0.3$ pixels \cite{Grompone2010}. High precision aims at final 3D
reconstructions far more precise than this. So the measuring tool
error should be also far smaller, if only possible.

The obvious advantage of ``real'' plumb lines on 2D patterns with
straight lines on them, is that it is much easier to ensure a very
precise physical straightness for lines than a very precise physical
flatness for a physical pattern plate. Yet, the precision of the
resulting measurement or correction depends on the straightness of the
physical lines. In \cite{Brown1971}, to achieve a high precision,
plumb lines were made of very fine white thread and stabilized by
immersion of the plumb bobs in oil containers. Illumination was
provided by a pair of vertically mounted fluorescent fixtures. A dead
black background was provided for the plumb-line array to highlight
the contrast. The points on the lines were measured with a
microscope. The measuring process required from 5 to 6 hours for
generating 324 points. The digital procedure proposed here will be
automatic and faster.

For the applications where the precision is not a crucial point,
straight lines present in the natural scene can be used. In
\cite{Claus2005, Rosten2011}, the straight lines are obtained by
photographing the architectural scenes and points on the lines are
detected by Canny edge detector.

\begin{figure}[!htb]
\centering
\subfloat[The harp with a uniform opaque object as background]{
  \label{fig:harp_opaque_bg1}
  \includegraphics[width=0.46\textwidth]{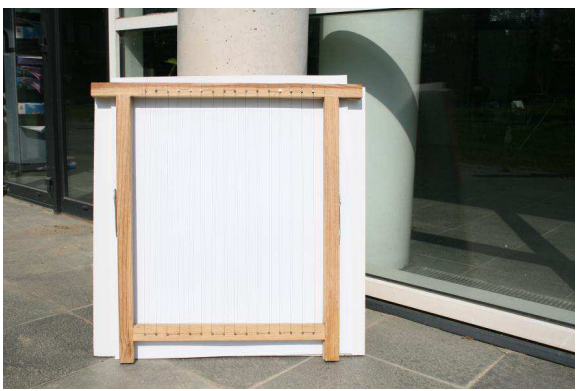}}
\quad
\subfloat[The harp with a translucent paper as background]{
  \label{fig:harp_translucent_bg1}
  \includegraphics[width=0.46\textwidth]{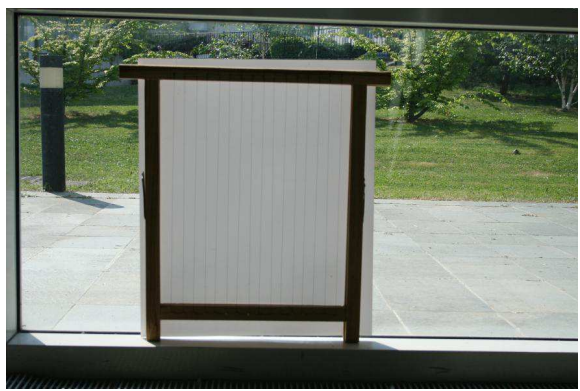}} \\
\subfloat[A close-up of the harp with a uniform opaque object as background]{
  \label{fig:closeup_harp_opaque_bg1}
  \includegraphics[width=0.46\textwidth]{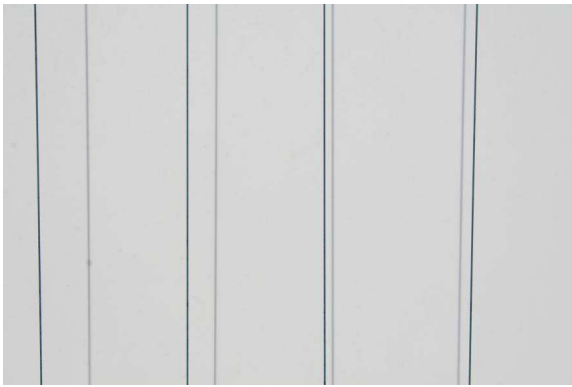}}
\quad
\subfloat[A close-up of the harp with a translucent paper as background]{
  \label{fig:closeup_harp_translucent_bg1}
  \includegraphics[width=0.46\textwidth]{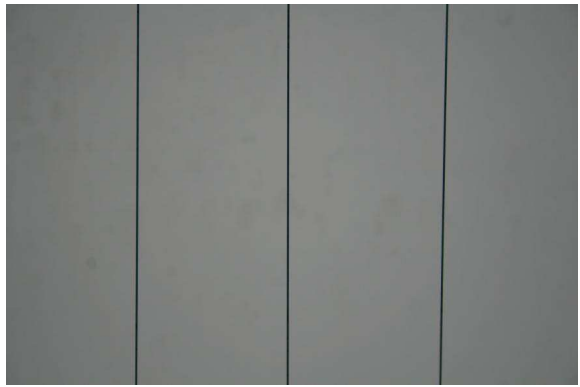}}
\quad
\caption{The ``calibration harp''. Shadows can be observed in (a) and
  (c), while there is no shadow in (b) or (d).}
\label{fig:harp1}
\end{figure}

In order to keep the high precision and simplify the fabrication
procedure, we built a simple calibration pattern by tightly stretching
strings on a frame as shown Fig.~\ref{fig:harp1}. The pattern looks
like the musical instrument, hence its name. The setup warrants the
physical straightness of the lines. Its construction does not require
any experimental skill, but only good quality strings. Indeed twisted
strings show local width oscillations; other strings do not have a
round section, and a little torsion also results in width variations
that can have a large spatial period. Rigid strings may have a
remanent curvature. Finally, a (tiny) gravity effect can be avoided by
using well stretched vertical lines.

\begin{figure}[!htb]
\centering
\subfloat[] {\label{fig:sewing_line1}
  \includegraphics[width=0.1\textwidth]{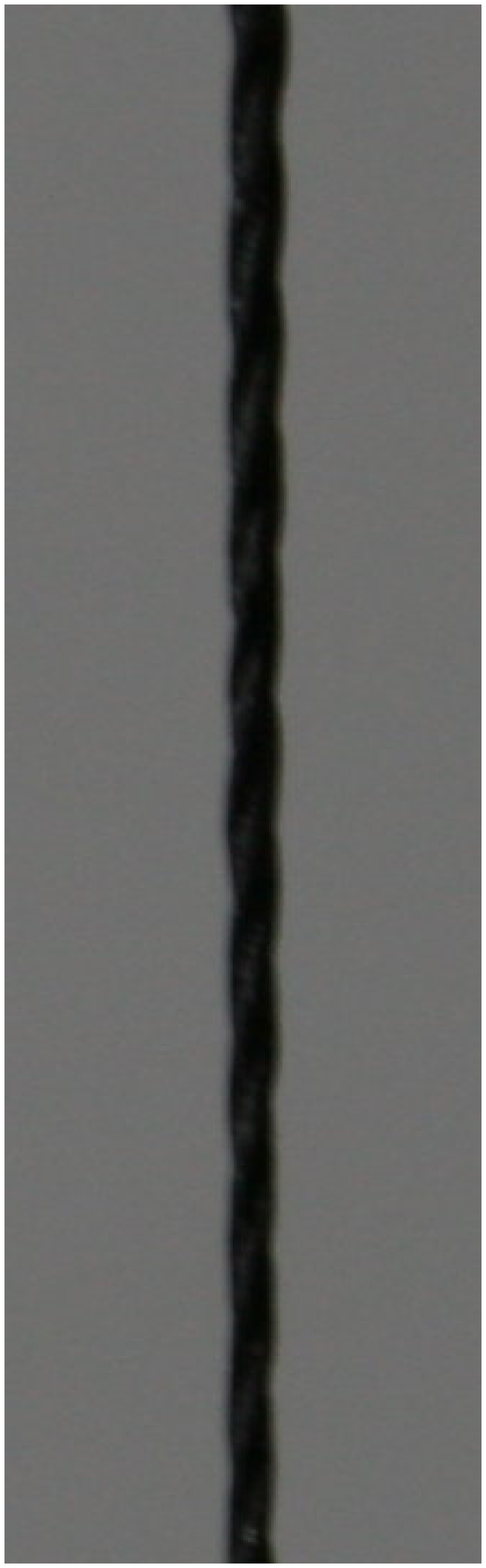}}
\hspace{20ex}
\subfloat[] {\label{fig:tennis_line1}
  \includegraphics[width=0.1\textwidth]{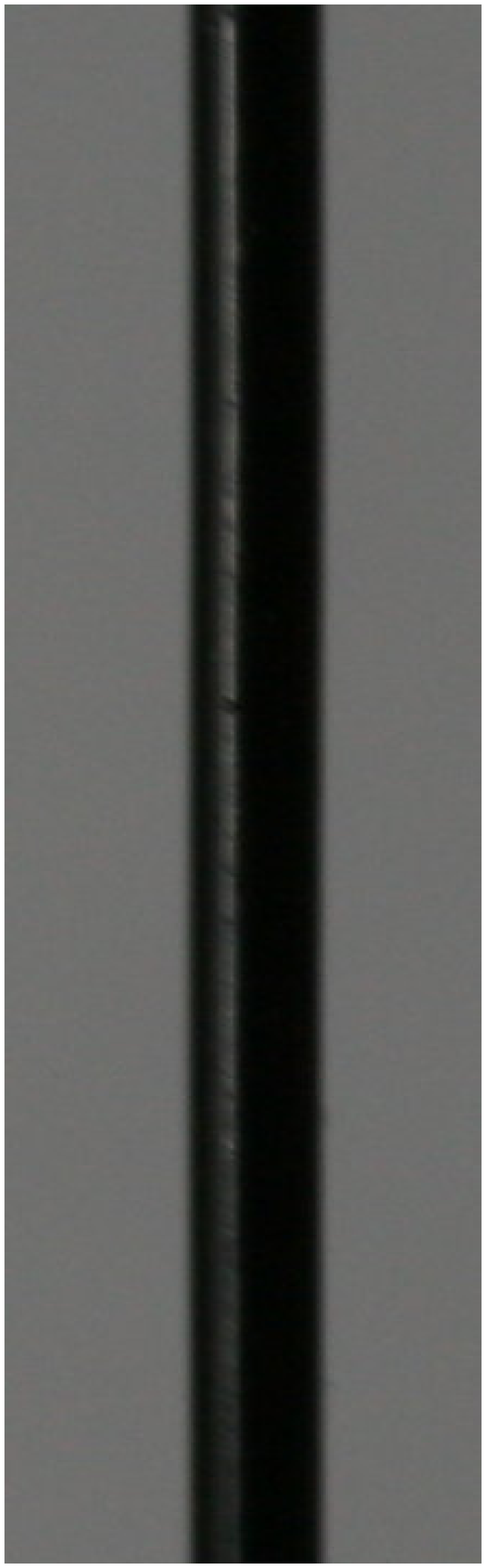}}
\hspace{20ex}
\subfloat[]{\label{fig:fishing_line1}
  \includegraphics[width=0.1\textwidth]{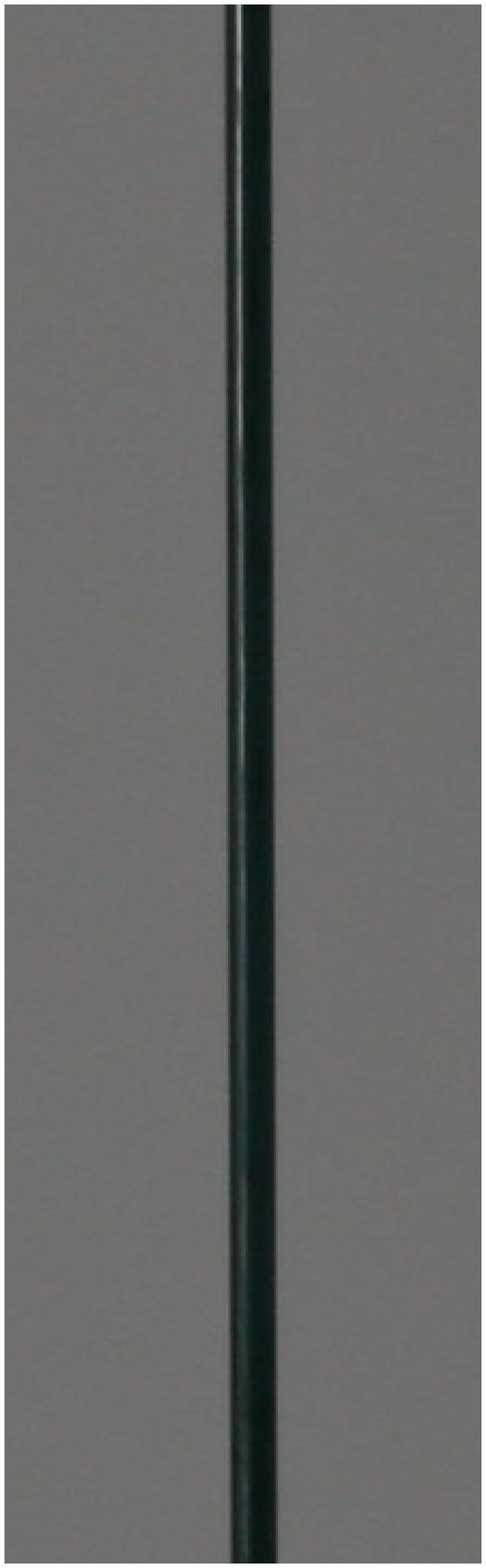}}
\caption{Three types of strings.
  (a) sewing line.
  (b) tennis racket line.
  (c) opaque fishing line.}
\label{fig:three_types_lines1}
\end{figure}

\begin{figure}[!htb]
\centering
\includegraphics[width=0.6\textwidth]{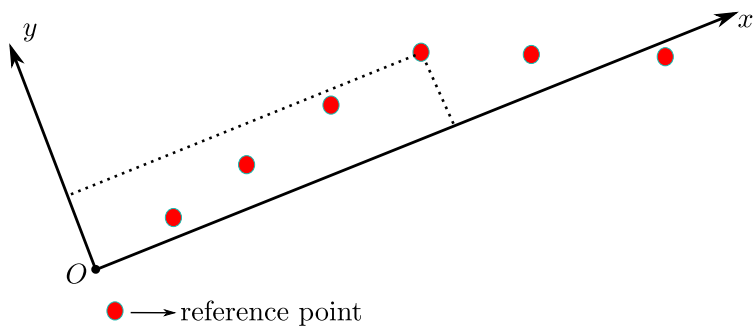}
\caption{The intrinsic coordinate system of the edge points extracted
  on the distorted line. The red points are the distorted edge
  points. The $x$ direction is determined by the direction of the
  regression line. The $x$-coordinate is the distance to the reference
  point along the regression line and the $y$-coordinate is the signed
  distance from the edge points to the regression line.}
 \label{fig:intrinsic_coord}
\end{figure}

\begin{figure}[!htb]
\centering
\subfloat[The sewing string]{
  \label{fig:sewing_string1}
  \includegraphics[width=0.3\textwidth]{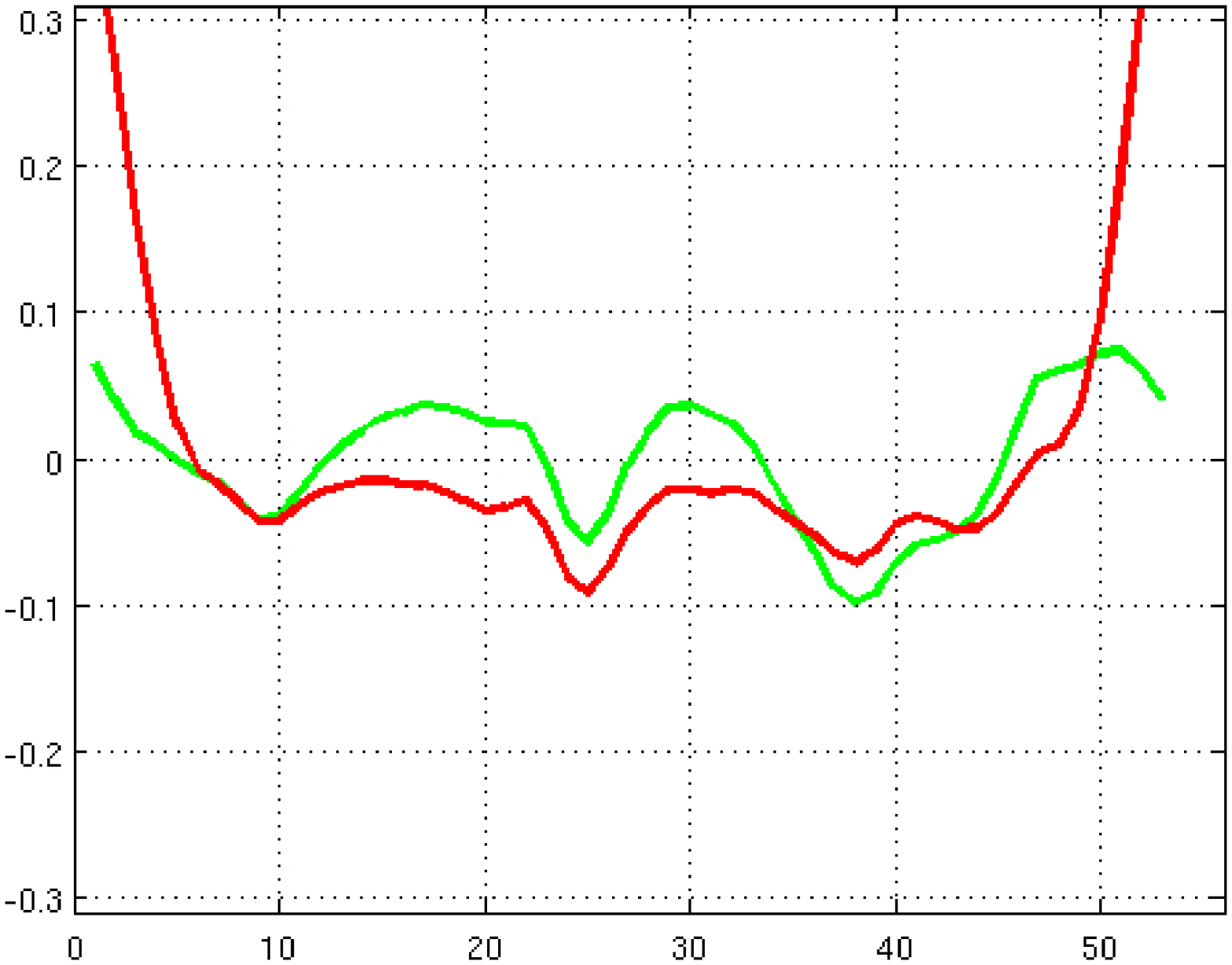}}
\subfloat[The tennis racket string]{
  \label{fig:tennis_string1}
  \includegraphics[width=0.3\textwidth]{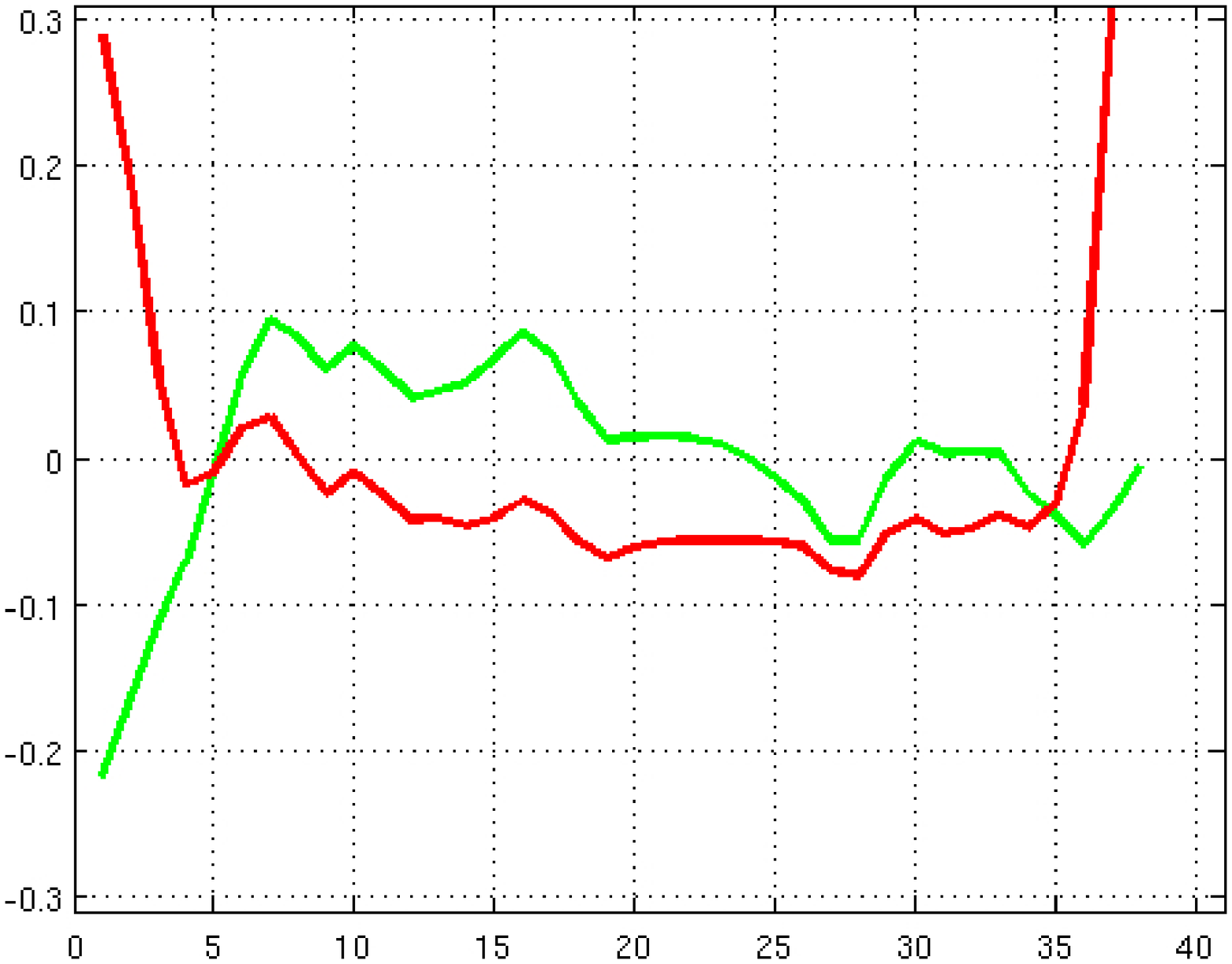}}
\subfloat[The opaque fishing string]{
  \label{fig:fishing_string1}
  \includegraphics[width=0.3\textwidth]{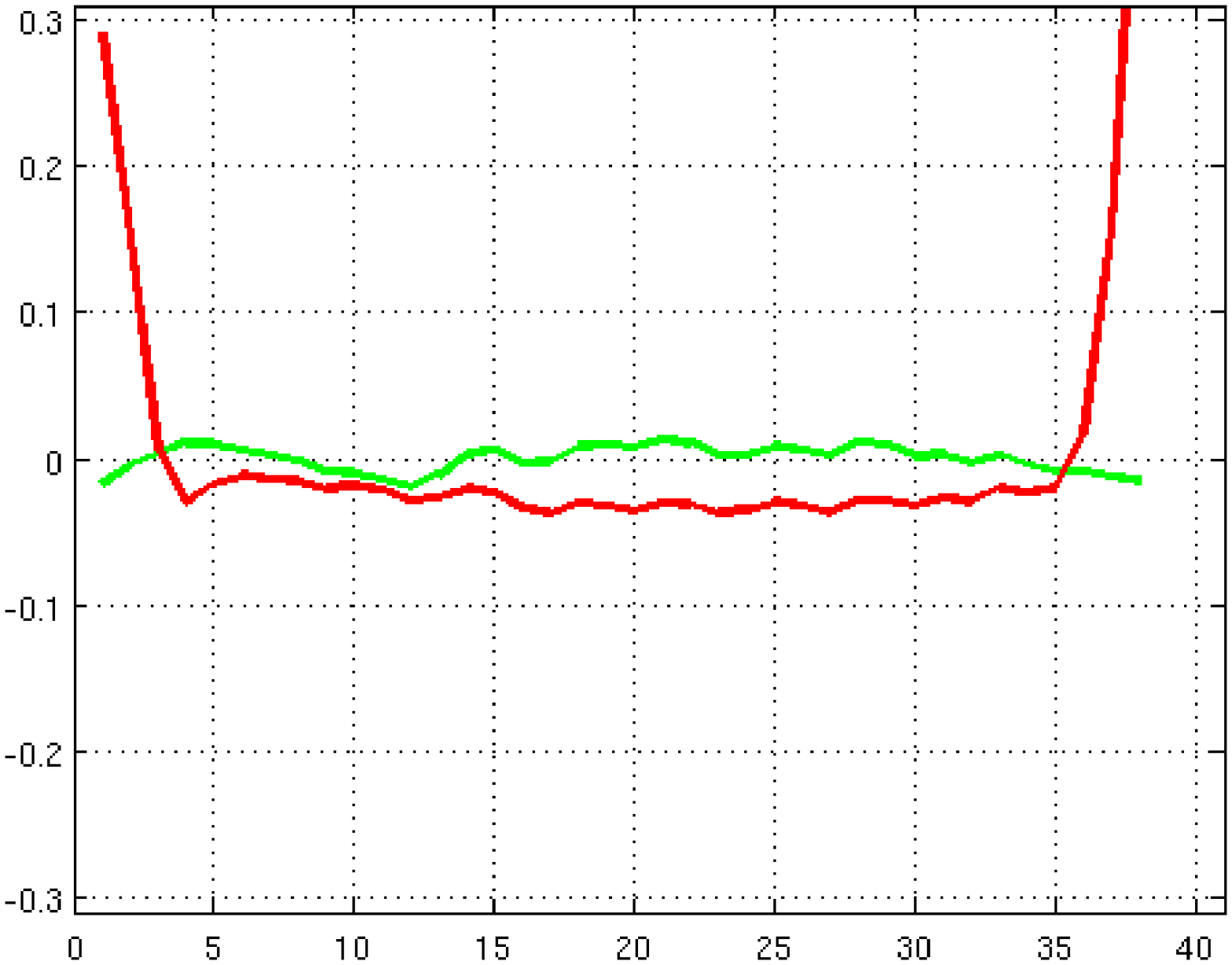}}
\caption{The small oscillation of the corrected lines is related to
  the quality of the strings. The green curves show the signed
  distance (in pixels) from the edge points of a corrected line to its
  regression line. The red curves show the high frequency of the
  corresponding distorted line. The corrected line inherits the
  oscillation from the corresponding distorted line.
  (a) The sewing string.
  (b) The tennis racket string.
  (c) The opaque fishing line. The $x$-axis is the index of edge
  points. The range of the $y$-axis is from $-0.3$ pixels to $0.3$
  pixels. The almost superimposing high frequency oscillation means
  that the high frequency of the distorted strings is not changed by
  the distortion correction. In such a case, the straightness error
  includes the high frequency of the distorted strings and does not
  really reflect the correction performance. So it is better to use a
  string which contains the smallest high frequency oscillation. Among
  the three types of strings, the opaque fishing string shows the
  smallest such oscillations. The larger oscillation of the sewing
  string is due to a variation of the thickness related to its twisted
  structure, while the tennis racket string is simply too rigid to be
  stretched, even though this is not apparent in
  Fig.~\ref{fig:tennis_line1}.}
\label{fig:string_high_freq1}
\end{figure}

In our experiments three different strings were tested: a sewing
string, a smooth tennis racket string, and an opaque fishing string,
all shown in Fig.~\ref{fig:three_types_lines1}. Sewing strings have a
braid pattern and their thickness oscillates. Tennis racket strings
are rigid and require a very strong tension to become
straight. Fishing strings are both smooth and flexible, and can
therefore be easily tightened and become very straight. The
transparent ones, however, behave like a cylindrical lens, producing
multiple complex edges. Opaque fishing strings end up being the best
choice to build a calibration harp. Fig.~\ref{fig:string_high_freq1}
shows an evaluation of the obtained straightness. We took photos of
the three types of strings, and corrected their distortion. The green
curves show the signed distance from edge points of a corrected line
to its regression line. The red curve shows the high frequency
component of the corresponding distorted line. The high frequency is
computed as follows:
\begin{itemize}

\item Transform the edge points extracted from the distorted line into
  the intrinsic coordinate system, which is determined by the
  direction of the regression line computed from these points. In this
  coordinate system, the $x$-coordinate is the distance between the
  edge points and a reference edge point along the regression line and
  the $y$-coordinate is the signed distance from the edge points to
  the regression line. This produces a one-dimensional signal (see
  Fig.~\ref{fig:intrinsic_coord}).

\item Apply a big Gaussian convolution of standard deviation $\sigma =
  40$ pixels on this one-dimensional signal to keep only the low
  frequency component.

\item The difference between this convolved signal and the original
  signal in the intrinsic coordinate system is considered as the high
  frequency. The red curves in Fig.~\ref{fig:string_high_freq1} show
  the high frequency (due to the border effect of Gaussian
  convolution, there is a sharp increase of magnitude at the border).

\end{itemize}

\begin{figure}[!htb]
\centering
\subfloat[Photo of the harp taken against the sky]{
  \label{fig:sky_bg1}
  \includegraphics[width=0.46\textwidth]{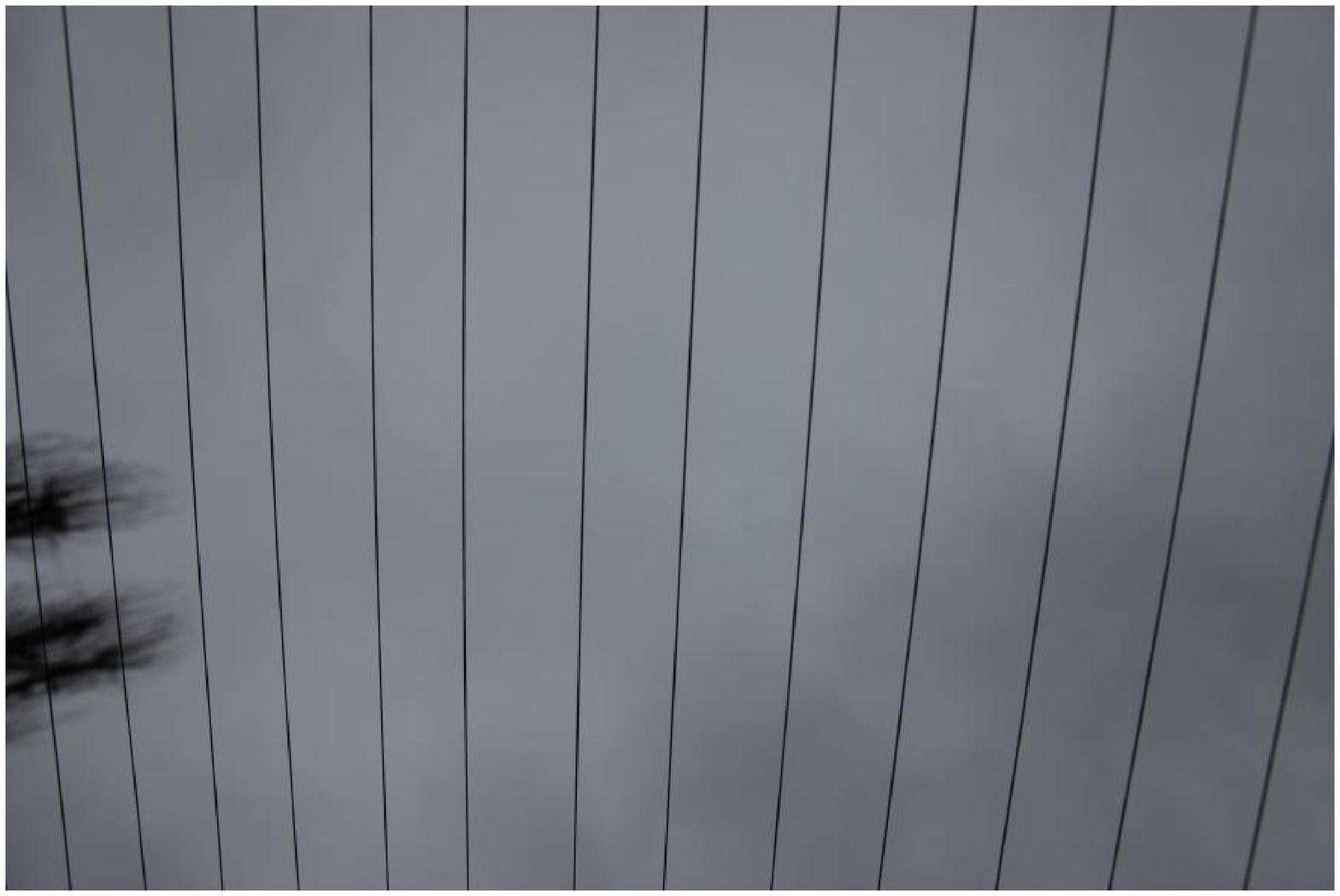}} \quad
\subfloat[Photo of the harp taken against a translucent paper using a
  tripod]{
  \label{fig:paper_bg1}\includegraphics[width=0.46\textwidth]{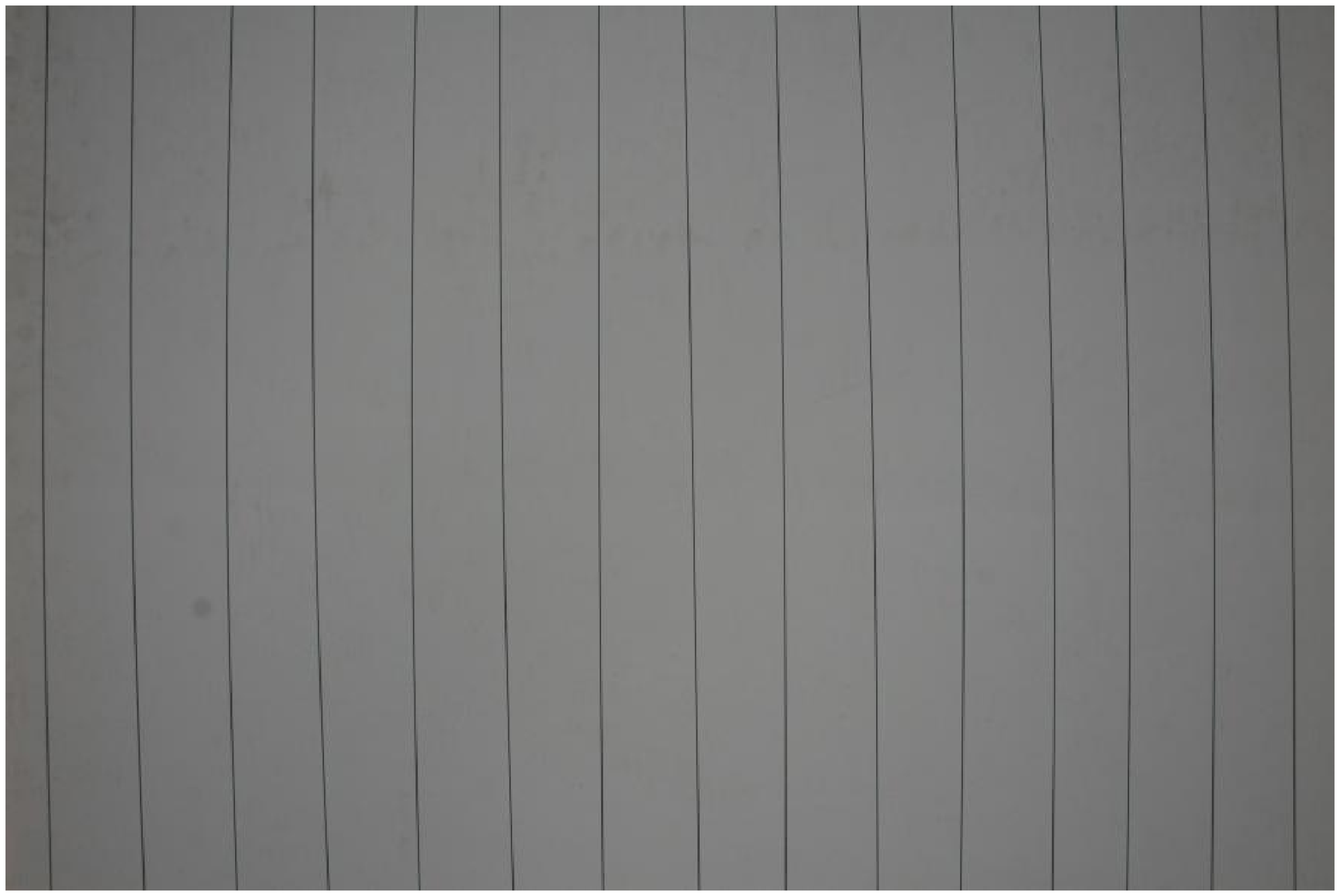}}
\caption{The quality of photos depends on the harp, its background and
  also the stability of camera for taking photos.}
\label{fig:background1}
\end{figure}

To ensure the precision of the edge detection in the string images, a
uniform background whose color contrasts well with the string color
must be preferred. Using an opaque background is not a good idea
because this requires a direct lighting and the strings project
shadows on the background (Fig.~\ref{fig:harp_opaque_bg1} and
\ref{fig:closeup_harp_opaque_bg1}). The sky itself is hardly usable: a
large open space is needed to avoid buildings and trees entering the
camera field of view, and clouds render it inhomogeneous, see
Fig.~\ref{fig:background1}. The simplest solution is to place a
translucent homogeneous paper or plastic sheet behind the harp and to
use back illumination, preferably natural light to make it more
uniform (see Fig.~\ref{fig:harp_translucent_bg1} and
\ref{fig:closeup_harp_translucent_bg1}).

The acquisition aspects are also important for producing high quality
measurements: lens blur, motion blur, aliasing, noise, must be as
reduced as physically possible. To that aim a tripod and timer were
used to reduce camera motions, but also to avoid out of focus strings
while taking photos at different orientations. Of course, changing
focus changes the distortion. Thus each distortion calibration must be
done for a fixed focus, and is associated with this focus.

\section{Straight Edges Extraction}\label{sec:edge_detection}

In this section we describe the procedure to extract accurately and
smooth the aligned edge points, which will be used to measure the
distortion.

Devernay's algorithm \cite{Devernay1995} is the classic sub-pixel
accurate edge detector. The implementation of Devernay's detector is
very simple since it is derived from the well-known Non-Maxima
Suppression method \cite{Canny1986, Deriche1987}. On good quality
images (SNR larger than 100), Devernay's detector can attain a
precision of about $0.05$ pixels.

Straightness measurements require the detection of groups of edge
points that belong to the same physical straight line, and the
rejection of points that do not belong to any line. To this aim, line
segments are detected on the image using the LSD algorithm
\cite{Grompone2010lsd, IPOLLSDpaper}. When applied to photographs of
the calibration harp, the detection essentially corresponds to the
strings. In case of a strong distortion, one string edge could be cut
into several line segments.

\begin{figure}[!htb]
\centering
\includegraphics[width=0.7\textwidth]{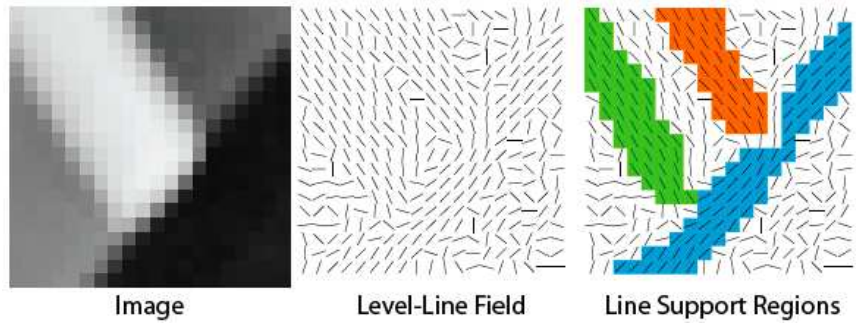}
\caption{The LSD algorithm computes the level-line field of the image.
  The level line field defines at each pixel the direction of the
  level line passing by this pixel. The image is then partitioned into
  connected groups that share roughly the same level-line
  direction. They are called \emph{line support regions}. Only the
  validated regions are detected as line segments. Devernay's edge
  points belonging to the same validated \emph{line support region}
  are considered as the edge points of the corresponding line
  segment.}
\label{fig:line-support-regions}
\end{figure}

LSD works by grouping connected pixels into \emph{line support
  regions}, see Fig.~\ref{fig:line-support-regions}; for more details
we refer to \cite{Grompone2010lsd, IPOLLSDpaper}. These regions are
then approximated by a rectangle and validated. The \emph{line support
  region} links a line segment to its support pixels. Thus, Devernay's
edge points that belong to the same \emph{line support region} can be
grouped as aligned; points belonging to none are ignored.

For photos of strings, almost every pixel along each side of one
string is detected as an edge point at sub-pixel precision. So there
are about $1000$ edge points detected for a line of length of about
$1000$ pixels. This large number of edge points opens the possibility
to further reduce the detection and aliasing noise left by
sub-sampling the edge points.

\begin{figure}
\centering
\includegraphics[width=0.7\textwidth]{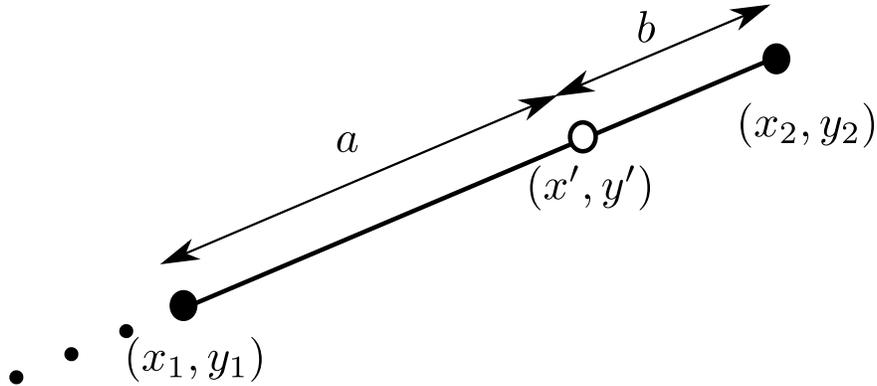}
\caption{Line re-sampling. The black dots $(x_1, y_1)$, $(x_2, y_2)$,
  $\ldots$ are the edge points extracted by Devernay's detector. They
  are irregularly sampled along the line. The re-sampling (in white
  dots) is made along the line with the uniform length step
  $d$. Linear interpolation is used to compute the re-sampled points.}
\label{fig:resampling}
\end{figure}

The sub-sampling step must be done carefully. First the edge points
are re-sampled to warrant a uniform sampling step along the edge; this
will facilitate the following steps. The re-sampling uses a step of $d
= L/N$ where $L$ is the length of a line and $N$ is the number of
extracted edge points on the line. The interpolation of an edge point
$(x', y')$ between two adjacent edge points $(x_1, y_1)$ and $(x_2,
y_2)$ is expressed by
\begin{eqnarray*}
  x' & = & \frac{b}{a+b} x_1 + \frac{a}{a+b} x_2, \\
  y' & = & \frac{b}{a+b} y_1 + \frac{a}{a+b} y_2,
\end{eqnarray*}
where $a$ and $b$ are the distances between the points, see
Fig.~\ref{fig:resampling}. Then, a Gaussian blur with $\sigma = 0.8
\times \sqrt{t^2 -1}$ is needed before a sub-sampling of factor $t$ to
avoid aliasing \cite{Morel2011sift}. We have two one-dimensional
signals ($x$-coordinate and $y$-coordinate of edge points) along the
length of the line. The Gaussian convolution is performed on both
one-dimension signals, parameterized by the length along the
edge. Finally, the sub-sampling of integer factor $t$ keeps one edge
point out of $t$.

\section{Distortion Measurements}\label{sec:measurements}

This section examines two natural distortion measurements that are
somewhat complementary.

\subsection{Root-Mean-Square Distance}

\begin{figure}
\centering
\includegraphics[width=0.8\textwidth]{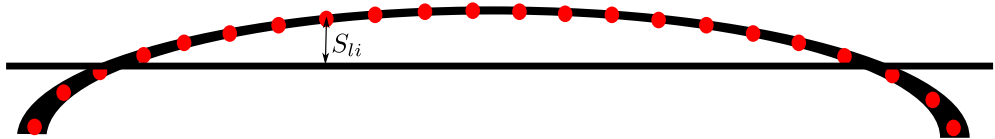}
\caption{The distance from a set of points to their global linear
  regression line.}
\label{fig:straightness}
\end{figure}

According to Theorem~\ref{theo:fundamental_prop}, the most direct
measure should be the straightness error, defined as the
root-mean-square (RMS) distance from a set of distorted edge points
that correspond to the same physical line, to their global linear
regression line, see Fig.~\ref{fig:straightness}.

Given $N$ edge points $(x_1,y_1),\ldots,(x_N,y_N)$ of a distorted
line, the regression line
\begin{equation}\label{eq:regression_line1}
  \alpha x + \beta y - \gamma = 0
\end{equation}
is computed by
\begin{equation*}
\alpha = \sin\theta, \quad \beta  = \cos\theta, \quad
\gamma = A_x \sin \theta + A_y \cos \theta,
\end{equation*}
where
\begin{gather*}
A_x = \frac{1}{N}\sum_{i=1}^N x_i, \quad
A_y = \frac{1}{N}\sum_{i=1}^N y_i, \\
V_{xx} = \frac{1}{N}\sum_{i=1}^N (x_i-A_x)^2, \,
V_{xy} = \frac{1}{N}\sum_{i=1}^N (x_i-A_x)(y_i-A_y), \,
V_{yy} = \frac{1}{N}\sum_{i=1}^N (y_i-A_y)^2, \\
\tan 2\theta = -\frac{2V_{xy}}{V_{xx}-V_{yy}}.
\end{gather*}
Since $(\alpha,\beta)$ is a unit vector, the signed distance from
point $(x_i,y_i)$ to the line is given by
$$
   S_i = \alpha x_i + \beta y_i - \gamma.
$$
Given $L$ lines, with $N_l$ points in line $l$, the total sum of
squared signed distance is given by
\begin{equation}
  S = \sum_{l=1}^{L} \sum_{i=1}^{N_{l}} |S_{li}|^2 = \sum_{l=1}^{L} \sum_{i=1}^{N_{l}}
      (\alpha_l x_{li} + \beta_l y_{li} - \gamma_{l})^2.
\end{equation}
Thus, the RMS straightness error is defined as
\begin{equation}\label{eq:straightness}
  d = \sqrt{\frac{S}{N_T}}.
\end{equation}
where $N_T = \sum_{l=1}^{L}N_{l}$ is the total number of points.

\subsection{Maximal error}

An alternative measure is the average maximal error defined by
\begin{equation}\label{eq:d_max}
  d_{\text{max}} = \sqrt{\frac{\sum_{l=1}^L |\max_i S_{li} - \min_i S_{li}|^2}{L}}.
\end{equation}
In the classic camera maker practice, the maximal error is defined by
$$
   \max_l |\max_i S_{li} - \min_i S_{li}|,
$$
which would become instable with the calibration harp, some of the
strings being potentially distorted by blur or wrong detection.

\begin{figure}
\centering
\includegraphics[width=0.45\textwidth]{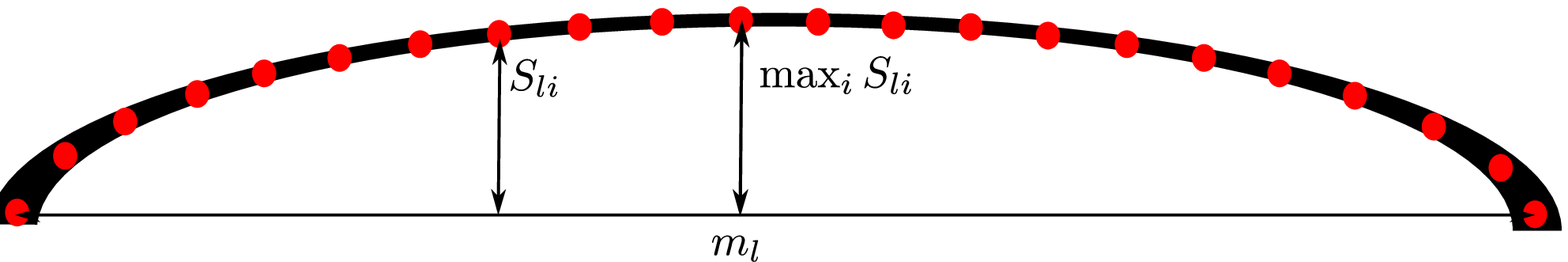}
\includegraphics[width=0.45\textwidth]{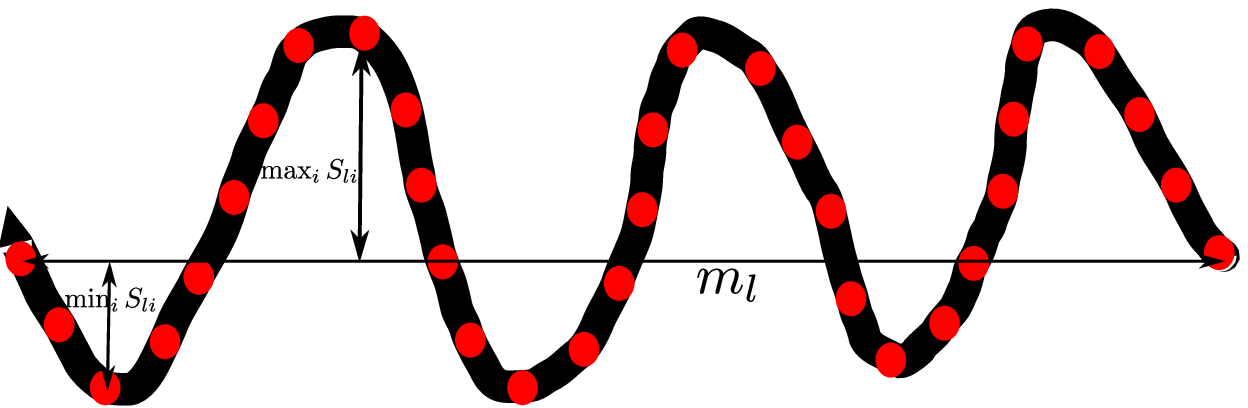}
\caption{Left: traditional distortion measure: the maximal distance to
  the line defined by the extremities of the edge. Right: The
  regression line crosses the distorted line; the difference between
  the maximal and minimal signed distance to the line measures the
  full width of the distorted line.}
\label{fig:arc}
\end{figure}

This measure is traditionally used in manual settings, for example see
Fig.~\ref{fig:dxo_distortion}. While traditionally the measures are
made relatively to the line joining the extremities of the distorted
edge, see Fig.~\ref{fig:arc}, here we use the signed distance to the
regression line to make it more comparable to the previous
measure. The use of a signed distance and the difference between the
maximal and minimal value is needed to handle correctly the fact that
there are values on both sides of the regression line, see
Fig.~\ref{fig:arc}.

\section{Applications}\label{sec:applications}

In this section, real photographs of the calibration harp will be used
to evaluate the residual camera distortion when this distortion has
been corrected with three state-of-the-art correction methods or two
popular software. In addition, we can feed any plumb-line method with
the precise edge points detected on the harp images to improve the
correction precision. The efficient Alvarez et~al. algorithm thus
becomes an automatic parametric distortion correction method. Our lens
distortion measurement algorithm can be tested on the online demo
version available at \url{http://bit.ly/lens-distortion}.

\subsection{Precision}

It is reported in \cite{Devernay1995} that Devernay edge points have a
precision better than $0.05$ pixel under zero-noise condition.  As
proposed in Section~\ref{sec:edge_detection}, the precision of
Devernay edge points can be further improved by applying Gaussian
convolution of standard deviation $0.8 \times \sqrt{t^2 -1}$, followed
by a sub-sampling of factor $t$. The only parameter to adjust here is
the factor $t$, which corresponds to the assumed regularity of the
lens distortion. We are only interested in realistic lens distortion,
which makes a straight line globally convex or concave. Thus local
edge oscillations due to noise can be harmlessly removed. In the
experiments, the value of $t = 30$ was chosen, which is enough to
remove the local oscillation while keeping the global distortion.

\subsection{Measuring the residual error after distortion}

\begin{figure}[!htb]
\centering
\includegraphics[width=0.19\textwidth]{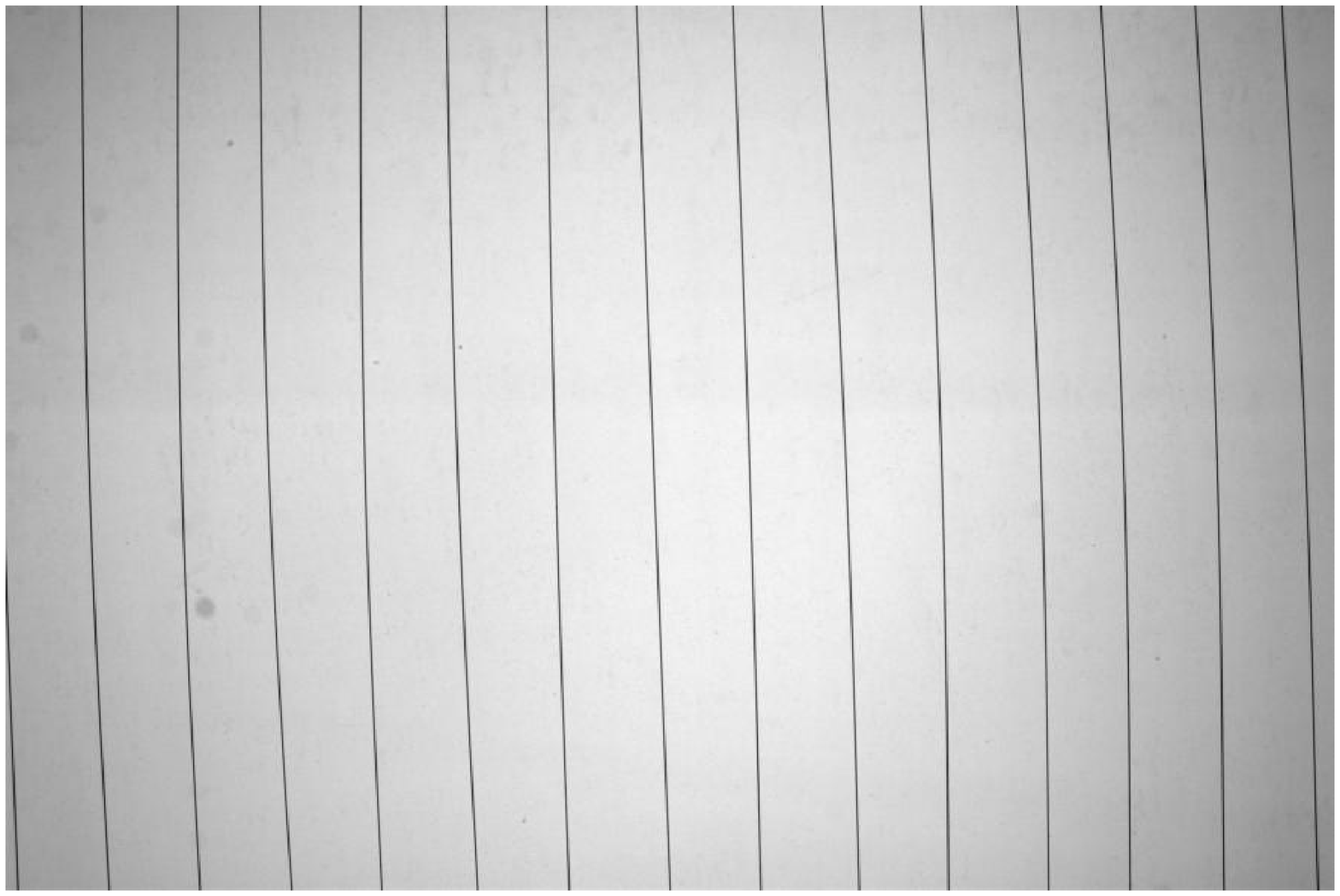}
\includegraphics[width=0.19\textwidth]{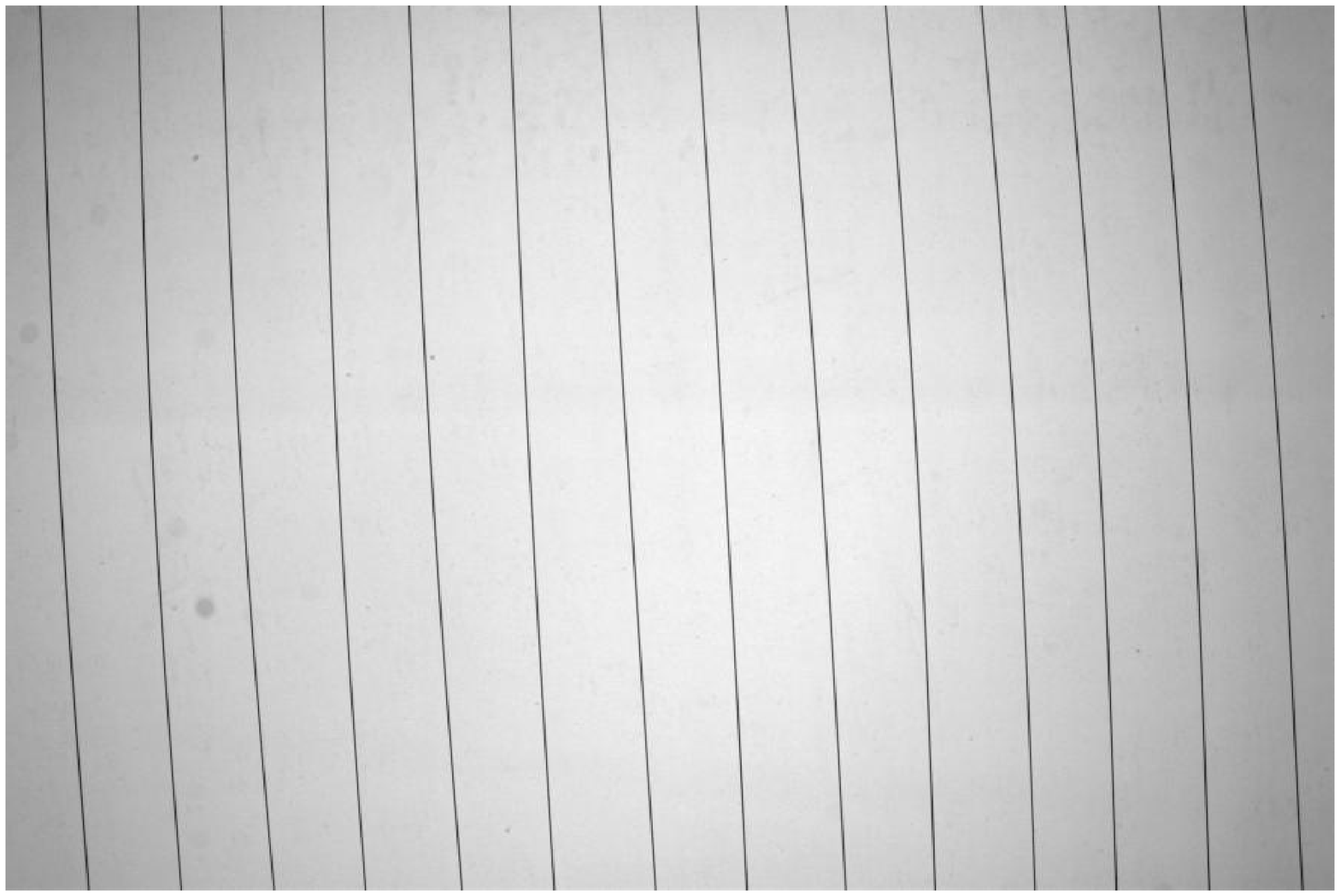}
\includegraphics[width=0.19\textwidth]{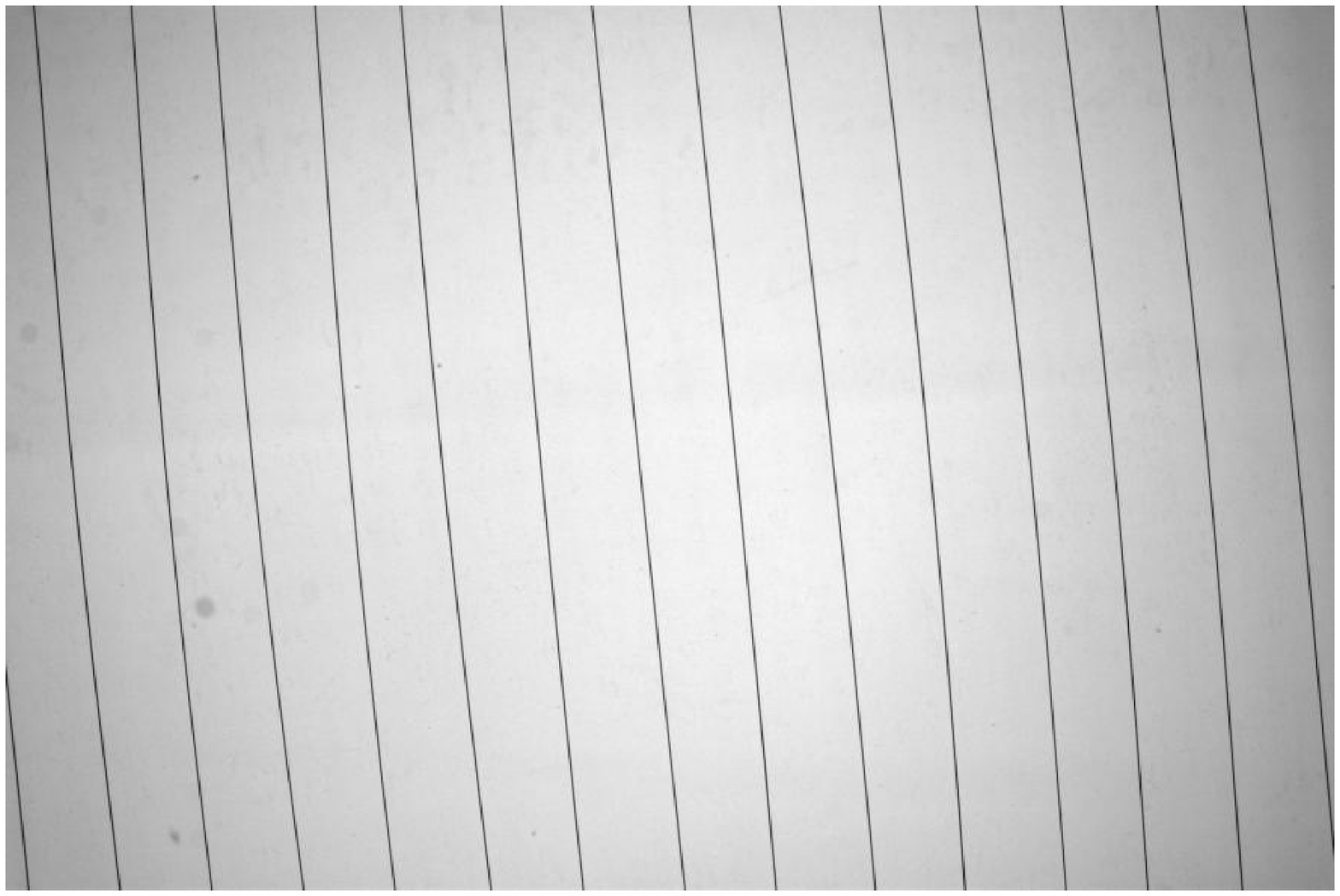}
\includegraphics[width=0.19\textwidth]{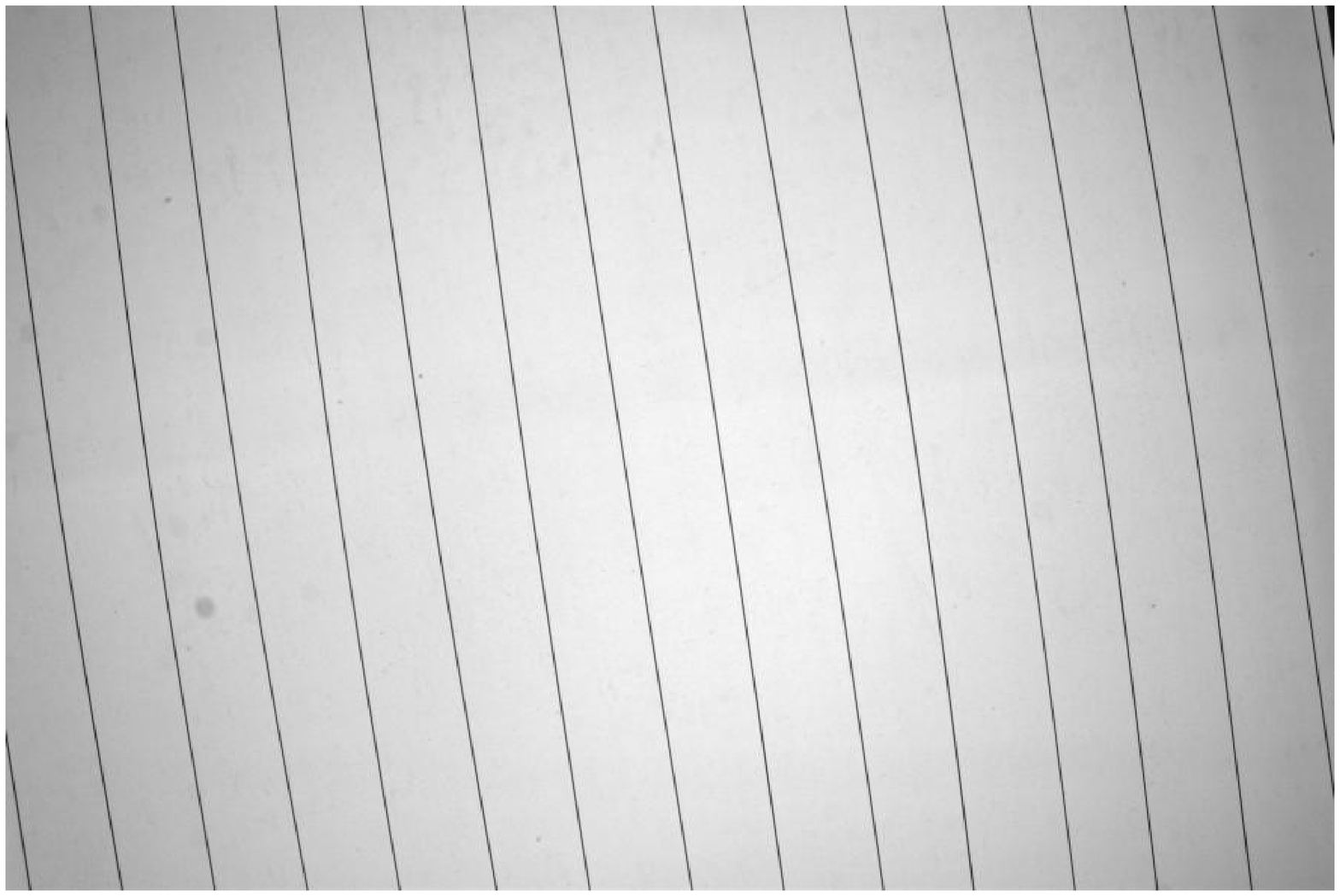}
\includegraphics[width=0.19\textwidth]{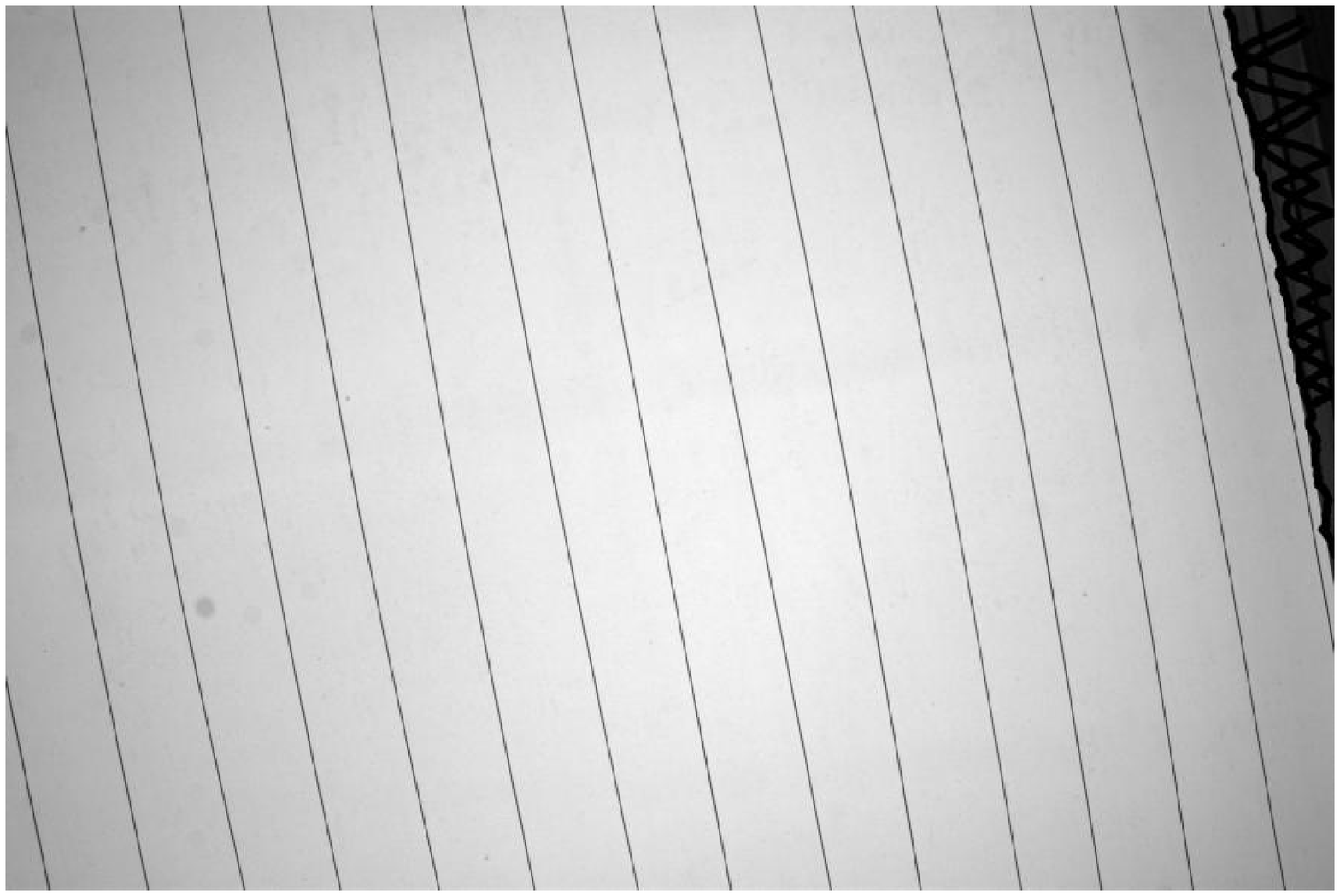} \\
\includegraphics[width=0.19\textwidth]{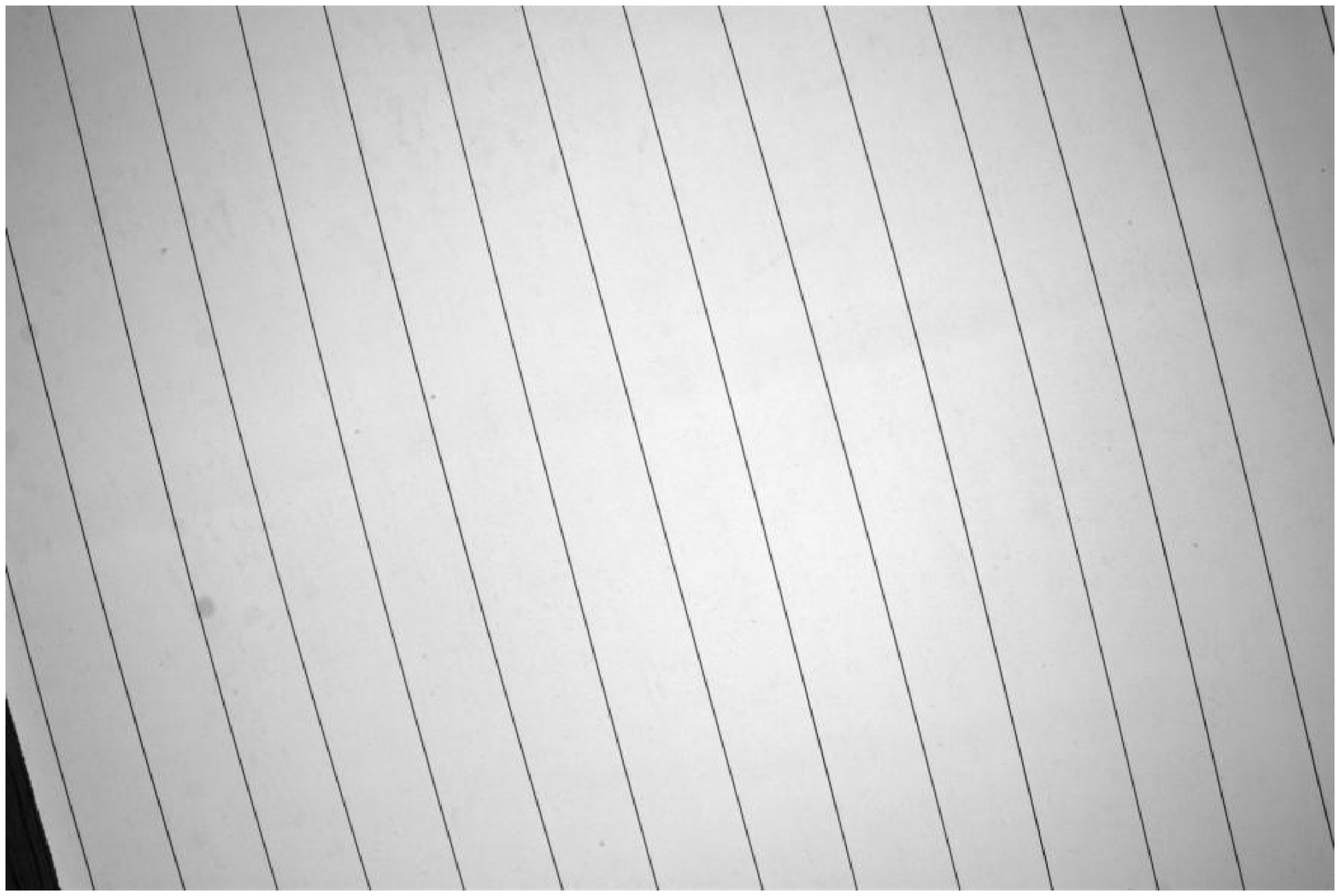}
\includegraphics[width=0.19\textwidth]{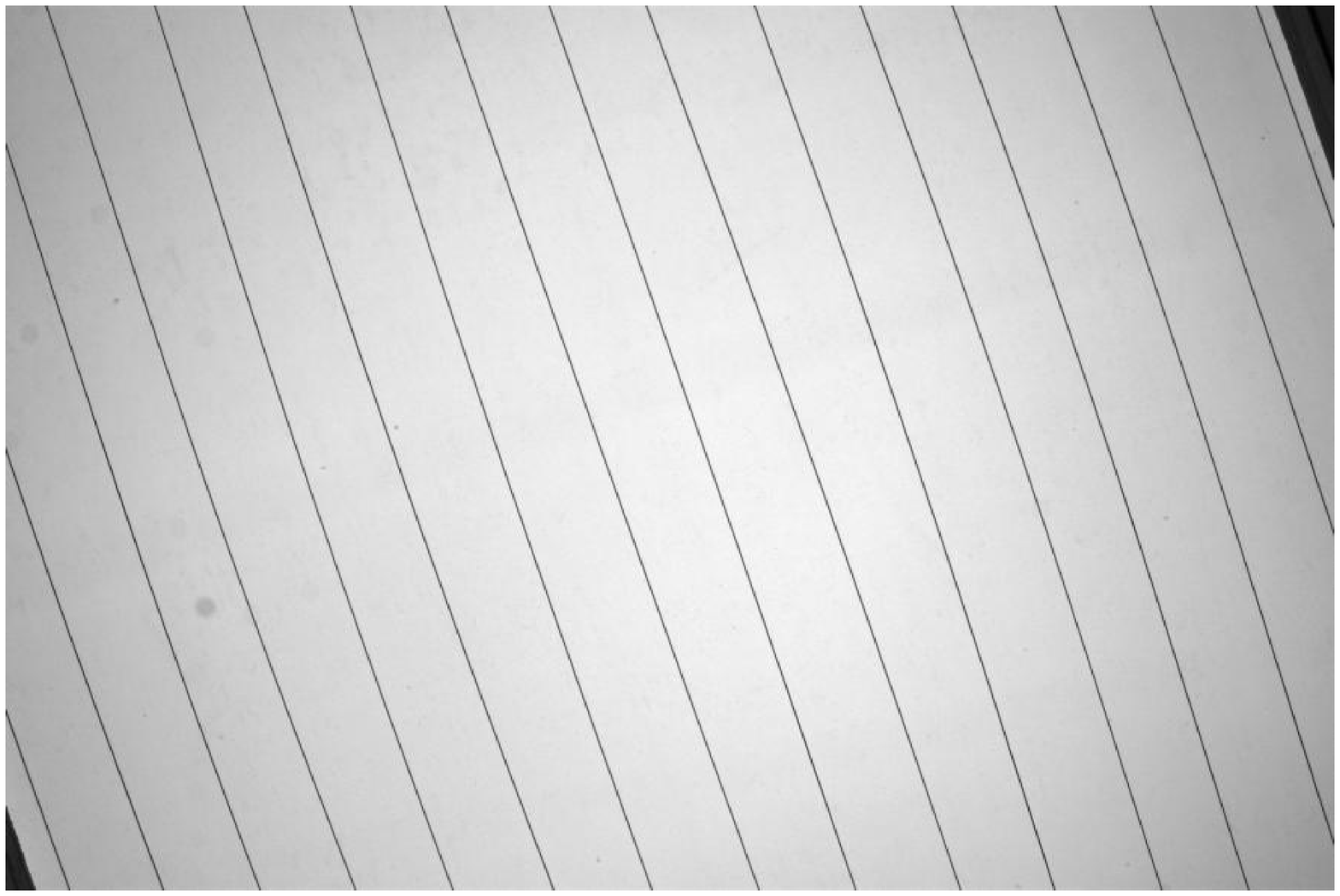}
\includegraphics[width=0.19\textwidth]{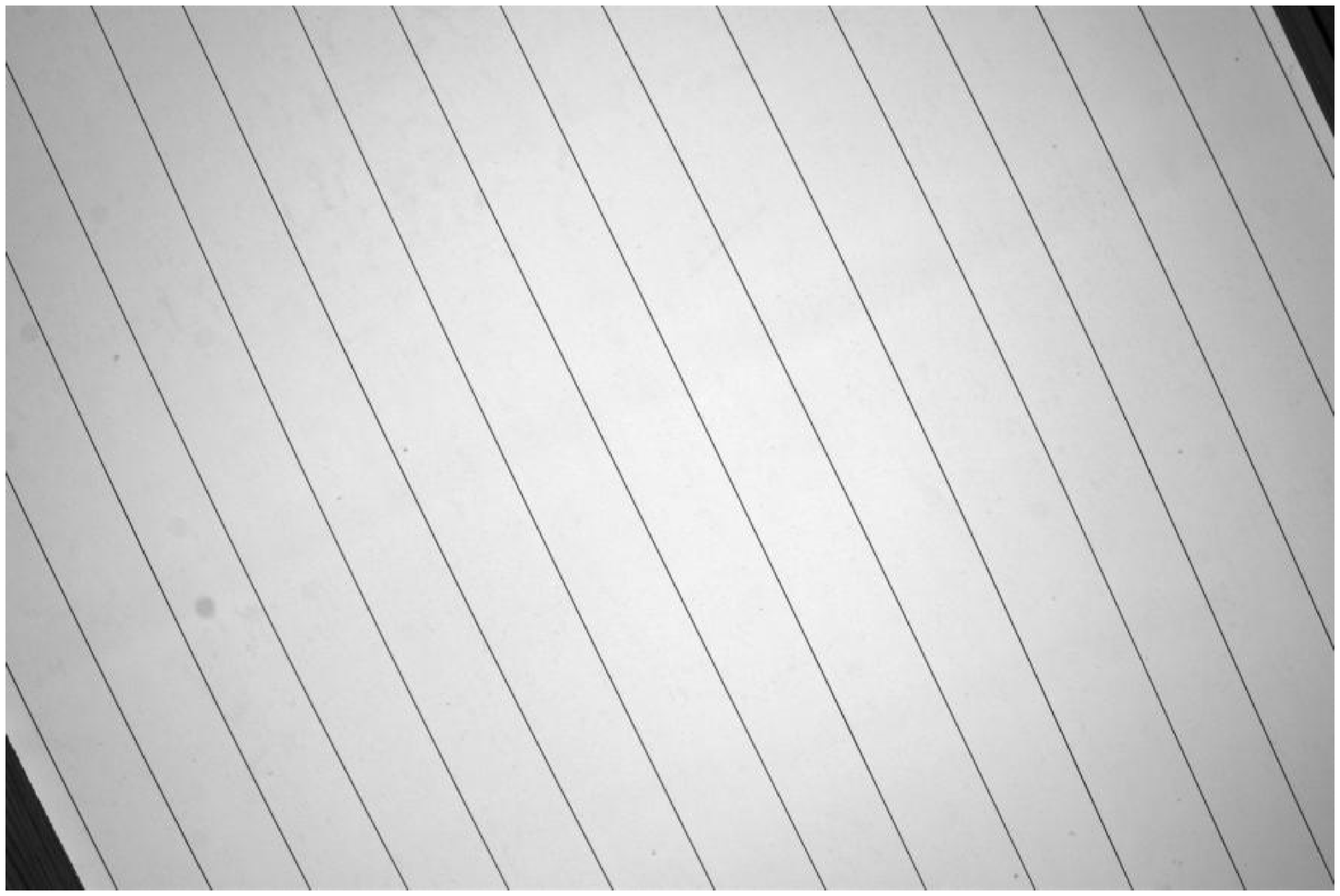}
\includegraphics[width=0.19\textwidth]{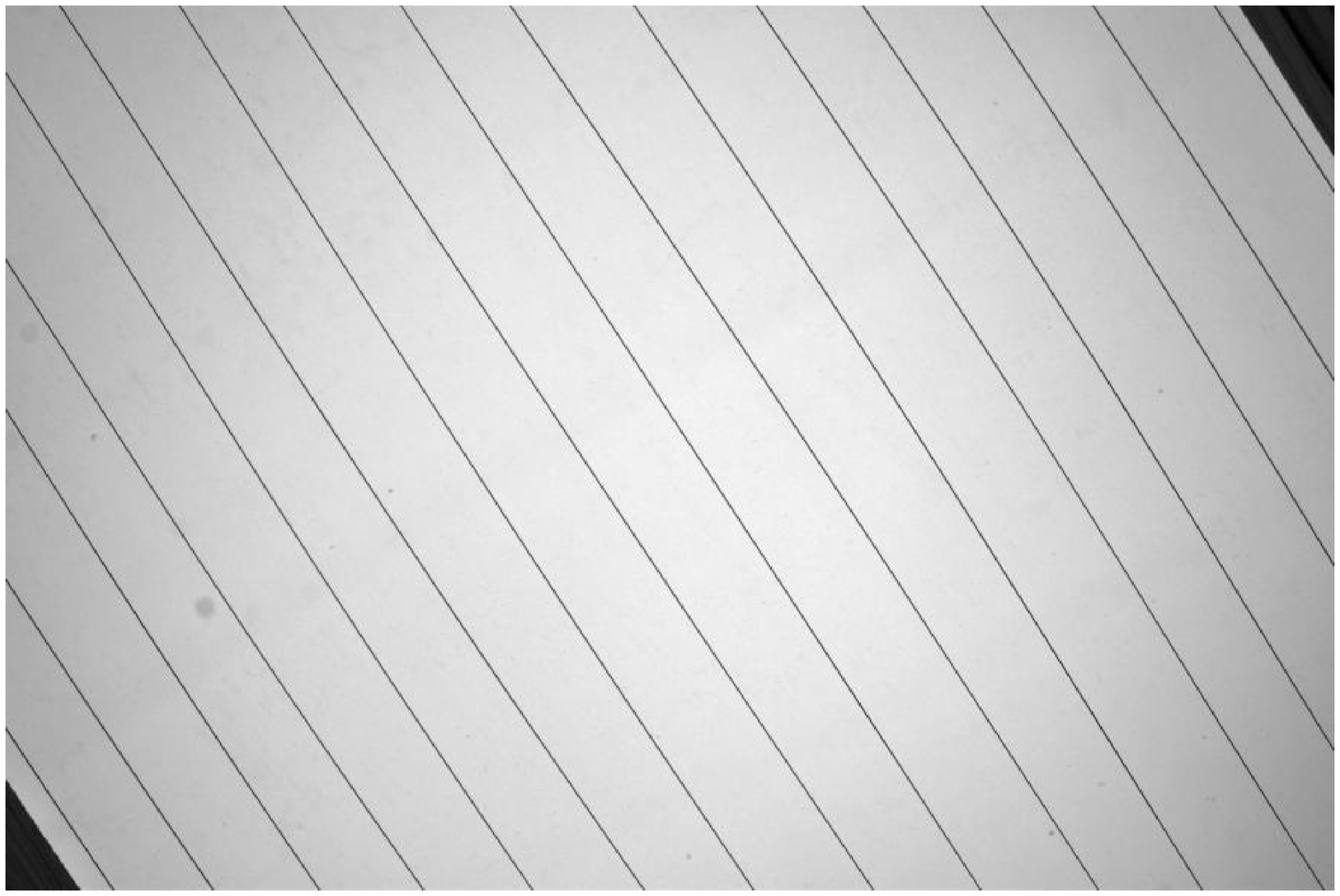}
\includegraphics[width=0.19\textwidth]{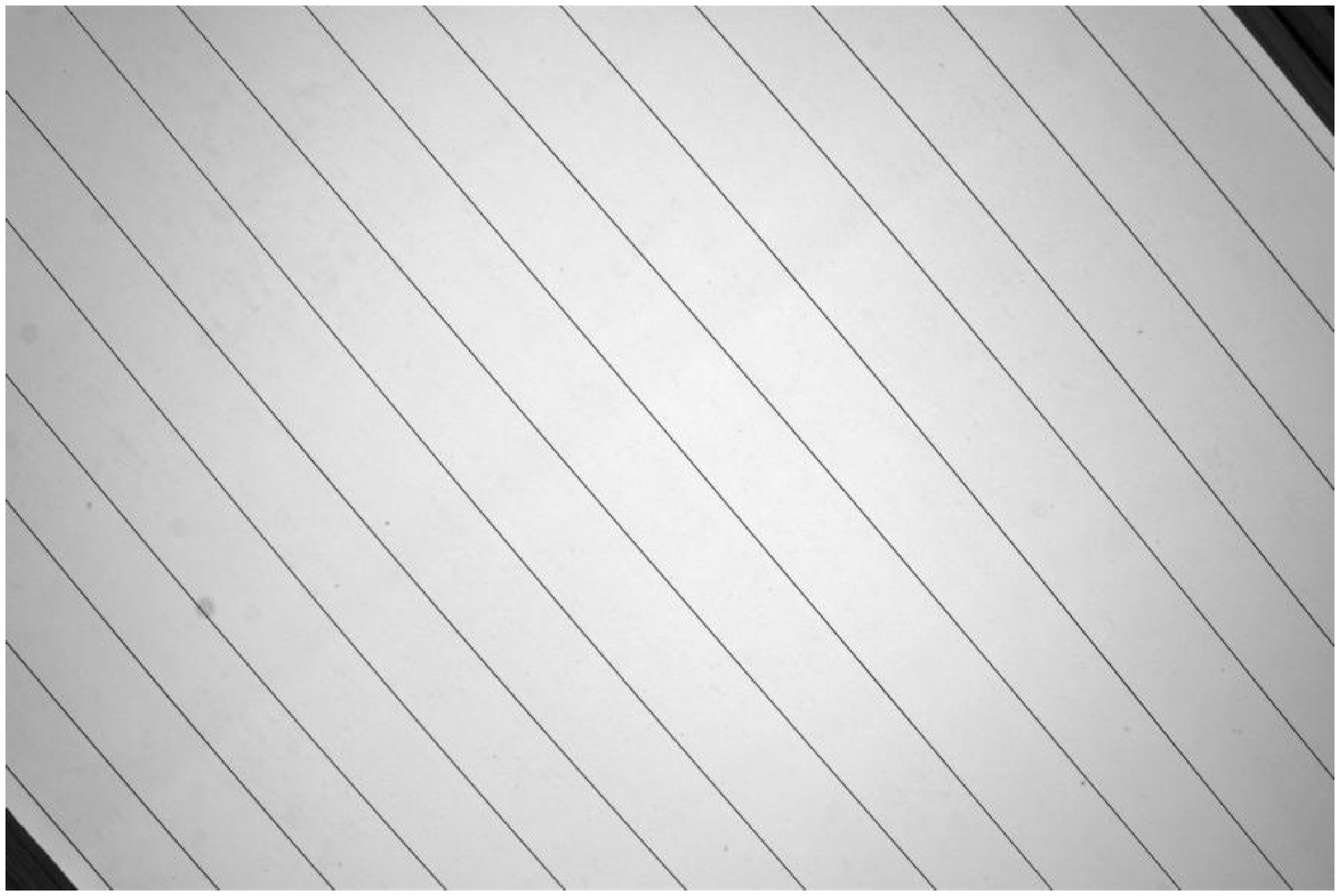}
\caption{The distorted photos of the ``calibration harp''.}
\label{fig:distorted_calib_harp_photos}
\end{figure}

As a first main application, the ``calibration harp'' permits to
evaluate the performance of any distortion correction algorithm by
measuring its residual distortion in corrected images. The procedure
is as follows:
\begin{enumerate}

  \item A series of photos of the ``calibration harp'' are taken at
    different orientations.

  \item These photos are processed by a camera distortion correction
    algorithm.

  \item The corrected images are normalized by a homography as
    described in section~\ref{sec:theorem}.

  \item The residual distortion is measured by the proposed method.

\end{enumerate}
Three distortion correction algorithms and two software were
tested. With the exception of the classic Lavest et~al. calibration
method, all the others are designed to only correct the lens
distortion without estimating the other camera parameters:
\begin{itemize}

  \item The Lavest et~al. method \cite{Lavest1998}: probably the most
    advanced pattern-based global camera calibration method, which
    estimates and corrects for the pattern non-flatness, using a
    bundle adjustment technique. Various distortion parameter
    configurations are allowed in this method: 2 radial parameters and
    2 tangential parameters for a partial distortion model; 2 radial
    parameters for a partial radial distortion model; 5 radial
    parameters for a complete radial distortion model; 5 radial
    parameters and 2 tangential parameters for a full distortion
    model.

  \item A non-parametric lens distortion correction method requiring a
    textured flat pattern \cite{Grompone2010}. The pattern is obtained
    by printing a textured image and pasting it on an aluminum plate,
    which is thick and solid.

  \item The DxO-Optics-Pro software: a program for professional
    photographers automatically correcting lens distortion (even from
    fisheyes), color fringing and vignetting, noise and blur. This
    software reads the EXIF of each image to know exactly what camera,
    lens and settings have been used. It therefore uses a fixed lens
    distortion estimation for each supported camera model.

   \item PTLens: Photoshop plug-in that corrects lens
     pincushion/barrel distortion, vignetting and chromatic
     aberration.

\end{itemize}
The distorted photographs to be corrected are shown in
Fig.~\ref{fig:distorted_calib_harp_photos} and
Table~\ref{tab:mesurements_on_different_algo} shows the residual
distortion measurements obtained by the calibration harp, after
applying the corrections specified by the various methods.

\begin{table}[!htb]
\caption{The distortion correction performance of three algorithms,
  measured by RMS distance $d$ and maximal distance $d_{\text{max}}$.}
\label{tab:mesurements_on_different_algo}
\begin{center}
\begin{tabular}{lccccccc}
\hline
method & $d$ (pixels) & $d_{\text{max}}$ \\ 
\hline
original distortion                           & $2.21$ & $6.70$ \\ 
\hline
Lavest (2 radial and 2 tangential parameters) & $0.07$ & $0.30$  \\ 
\hline
Lavest (2 radial parameters)                  & $0.07$ & $0.29$  \\ 
\hline
Lavest (full distortion parameters)           & $0.60$ & $3.00$  \\ 
\hline
Lavest (full radial distortion parameters)    & $0.59$ & $2.90$  \\ 
\hline
Textured pattern                              & $0.04$ & $0.16$  \\ 
\hline
DxO Optics Pro                                & $0.32$ & $0.99$  \\ 
\hline
PTLens                                        & $0.46$ & $1.51$  \\ 
\hline
\end{tabular}
\end{center}
\end{table}

The Lavest et~al. method depends on the parameter configuration of the
distortion model integrated in the global calibration process. Since
the global calibration process only minimizes the re-projection error
and does not control the distortion correction, it can happen that the
error in internal parameters compensates the error in external
parameters. In consequence, the minimized re-projection error is
small, but neither the estimated distortion parameters nor the other
parameters are correct. In fact this is the common drawback of global
camera calibration methods based on bundle adjustment. The textured
pattern based method requires a perfectly flat pattern. Even though it
is not very feasible to fabricate a perfectly flat pattern, a pattern
made of a thick and solid aluminium plate gives a good flatness
condition and thus a precise distortion correction. DxO Optics Pro
includes many pre-calibrated distortion models depending on the camera
type and parameters setting. But these distortion models are only
calibrated on several fixed focused distances and obtain by
interpolation the distortion models focused on the other
distances. Once the camera parameters are extracted from the EXIF of
each image, DxO Optics Pro asks the user to manually input the focused
distance before performing the correction. This makes the distortion
correction result less precise; considering this, the results are
surprisingly good. PTLens works in the similar manner as DxO Optics
Pro except that it does not ask users to provide the focused distance
information. It is not clear how PTLens recovers this information
which is not available in EXIF. Probably PTLens applies a fixed
correction for each focal length, independently of the focus. This
coarse approximation may explain why its correction precision is not
as good as DxO Optics Pro.

We also note that $d_\text{max}$ is always larger than $d$. This is
not surprising, since $d_\text{max}$ is the largest displacement with
respect to the linear regression line of the edge points.

\subsection{Strengthening Plumb-Line Distortion Correction Methods}

Any plumb-line distortion correction method requires as input the edge
points on distorted lines, which are themselves projections of 3D
straight lines. The distortion can then be corrected by aligning the
edge points belonging to the same line. To see this, let us introduce
the widely used radial distortion correction model:
\begin{eqnarray}
  x_u - x_c & = & f(r_d) ( x_d - x_c ) \\
  y_u - y_c & = & f(r_d) ( y_d - y_c )
\end{eqnarray}
with $(x_u, y_u)$ the corrected point, $(x_d, y_d)$ the distorted
point, $(x_c, y_c)$ the distortion center and $r_d = \sqrt{(x_d-x_c)^2
  + (y_d-y_c) ^2}$ the distorted radius. Usually $f(r_d)$ is a
polynomial of $r_d$ and can be written as:
\begin{equation}
  f(r_d) = k_0 + k_1 r + k_2 r^2 + \cdots + k_N r^N
\end{equation}
with $k_0$, $k_1$, $k_2$, $ \ldots$, $ k_N$ the distortion parameters.
Assume we have $L$ lines and there are $N_l$ points on line $l$,
$l=1,2,\ldots,L$. A natural way to correct the distortion is to
minimize the sum of squared distances from the corrected points to the
corresponding regression line:
\begin{equation} \label{eq:syn_lines_min}
 D = \frac{1}{L} \sum_{l=1}^{L} \frac{1}{N_l} \sum_{i=1}^{N_l} S_{li}^2
   = \frac{1}{L} \sum_{l=1}^{L} \frac{1}{N_l} \sum_{i=1}^{N_l}
     \frac{ (\alpha_l x_{u_{li}} + \beta_l y_{u_{li}} + \gamma_{l})^2 }%
     {\alpha_l^2 + \beta_l ^2}
\end{equation}
with the linear regression line $l: \alpha_l x + \beta_l y +
\gamma_{l} = 0$ computed from the corrected points $(x_{u_{li}},
y_{u_{li}}), i = 1, \ldots, N_l$. The only unknown parameters in $D$
are $k_0, k_1, \ldots, k_N$ and $(x_c, y_c)$. With an appropriate
initialization of these parameters, $D$ can be efficiently minimized
by non-linear optimization algorithms, like Levenberg-Marquardt
algorithm.

Instead of minimizing $D$, Alvarez et~al. \cite{Alvarez2009} proposed
to minimize the measurement:
\begin{equation}
        E = \frac{1}{L} \sum_{l=1}^L ( S_{xx}^l S_{yy}^l - (S_{xy}^l)^2  )
\end{equation}
where $S^l$ is the covariance matrix for the corrected points on the line $l$:
\begin{equation}
        S^l =   \begin{pmatrix}
                  S_{xx}^l & S_{xy}^l \\
                  S_{xy}^l & S_{yy}^l
                \end{pmatrix}
=
 \frac{1}{N_l}
 \begin{pmatrix}
        \sum_{i=1}^{N_l} (x_{u_{l,i}} - \overline{x_{u_{l,i}}})^2 & \sum_{i=1}^{N_l} (x_{u_{l,i}} - \overline{x}_{u_{l,i}})(y_{u_{l,i}} - \overline{y_{u_{l,i}}}) \\
        \sum_{i=1}^{N_l} (x_{u_{l,i}} - \overline{x_{u_{l,i}}})(y_{u_{l,i}} -\overline{y_{u_{l,i}}})  & \sum_{i=1}^{N_l} ( y_{u_{l,i}} - \overline{y_{u_{l,i}}} )^2
 \end{pmatrix},
\end{equation}
with $ \overline{x_{u_{l,i}}} = \frac{1}{{N_l}} \sum_{i=1}^{N_l}
x_{u_{l,i}}$ and $ \overline{y_{u_{l,i}}} = \frac{1}{{N_l}}
\sum_{i=1}^{N_l} y_{u_{l,i}}$. It can be proven \cite{Alvarez2009}
that this new energy function $E$ is always positive and equals to $0$
if and only if for each line its points are aligned. The functional
$E$ can be further written in the form of matrix-vector multiplication
\cite{Alvarez2009}:
\begin{equation}
E(\mathbf{k}) = \frac{1}{L} \sum_{l=1}^L \mathbf{k}^T A^l \mathbf{k} \mathbf{k}^T B^l \mathbf{k}
 - \mathbf{k}^T C^l \mathbf{k} \mathbf{k}^T C^l \mathbf{k}
\end{equation}
with $\mathbf{k} = (k_0, k_1, k_2, \ldots, k_N)^T$ the distortion
parameters in the form of vector and
\begin{eqnarray}
  A_{m,n}^l & = & \frac{1}{N_l} \sum_{i=1}^{N_l} ( (r_{d_{l,i}})^m x_{d_{l,i}}
    - \overline{ (r_{d_{l,i}})^m x_{d_{l,i}} } )
   ( (r_{d_{l,i}})^n x_{d_{l,i}} - \overline{ (r_{d_{l,i}})^n x_{d_{l,i}} } ) \\
  B_{m,n}^l & = & \frac{1}{N_l} \sum_{i=1}^{N_l} ( (r_{d_{l,i}})^m y_{d_{l,i}} - \overline{ (r_{d_{l,i}})^m y_{d_{l,i}} } )
 ( (r_{d_{l,i}})^n y_{d_{l,i}} - \overline{ (r_{d_{l,i}})^n y_{d_{l,i}} } ) \\
  C_{m,n}^l & = & \frac{1}{N_l} \sum_{i=1}^{N_l} ( (r_{d_{l,i}})^m x_{d_{l,i}} - \overline{ (r_{d_{l,i}})^m x_{d_{l,i}} } )
 ( (r_{d_{l,i}})^n y_{d_{l,i}} - \overline{ (r_{d_{l,i}})^n y_{d_{l,i}} } )
\end{eqnarray}
with $ \overline{ (r_{d_{l,i}})^m x_{d_{l,i}} } = \frac{1}{{N_l}}
\sum_{i=1}^{N_l} (r_{d_{l,i}})^m x_{d_{l,i}} $ and $ \overline{
  (r_{d_{l,i}})^m y_{d_{l,i}} } = \frac{1}{{N_l}} \sum_{i=1}^{N_l}
(r_{d_{l,i}})^m y_{d_{l,i}} $, $ m = 1, \ldots, N$ and $n = 1, \ldots,
N $. The trivial solution $\mathbf{k} = (0, 0, 0, \ldots, 0)^T$ can be
avoided by setting $k_0 = 1$.

\begin{figure}[!htb]
\centering
\subfloat[Natural image]{\label{subfig:natural_img}
  \includegraphics[width=0.31\textwidth]{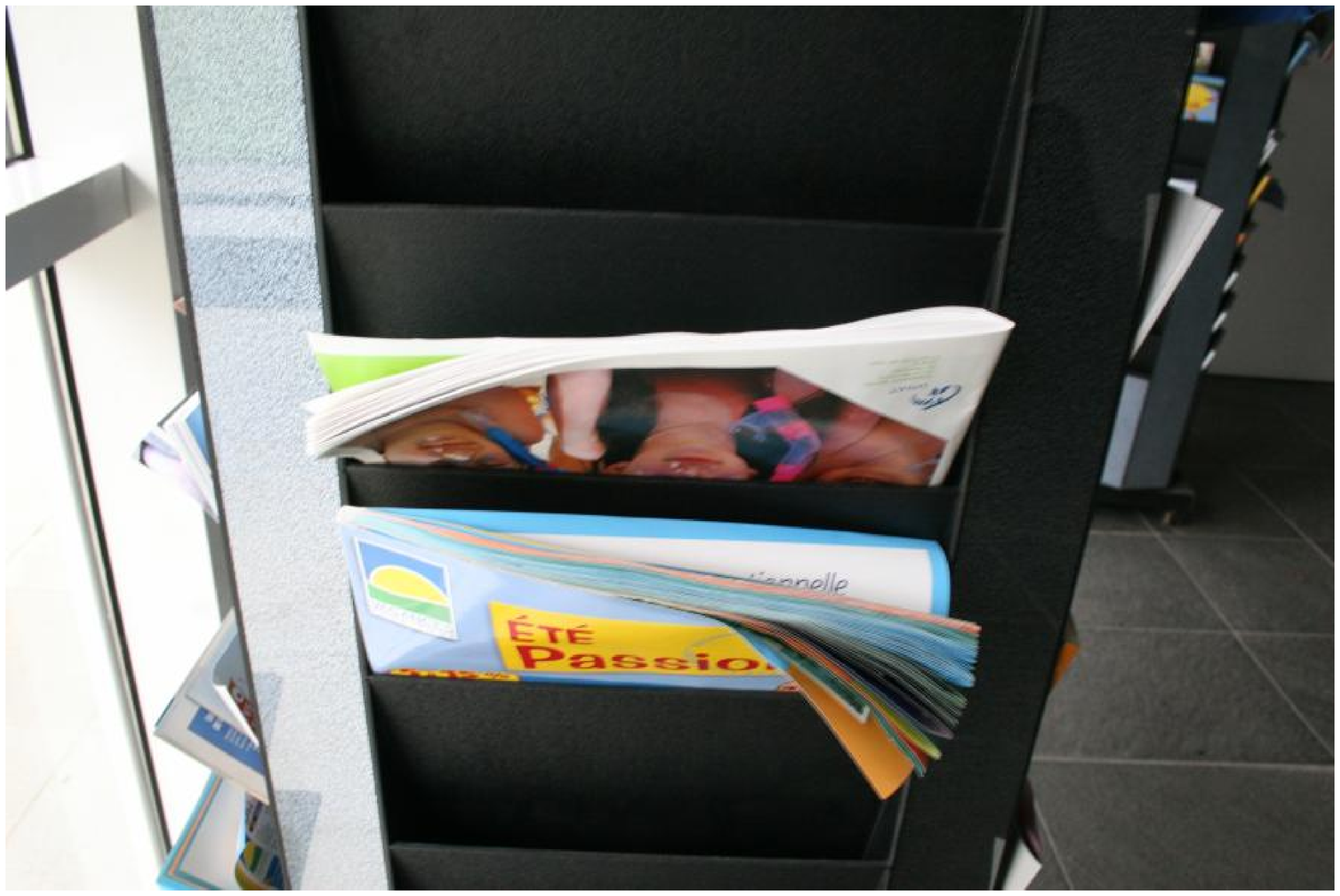}}
\subfloat[Grid pattern image]{\label{subfig:grid_img}
  \includegraphics[width=0.31\textwidth]{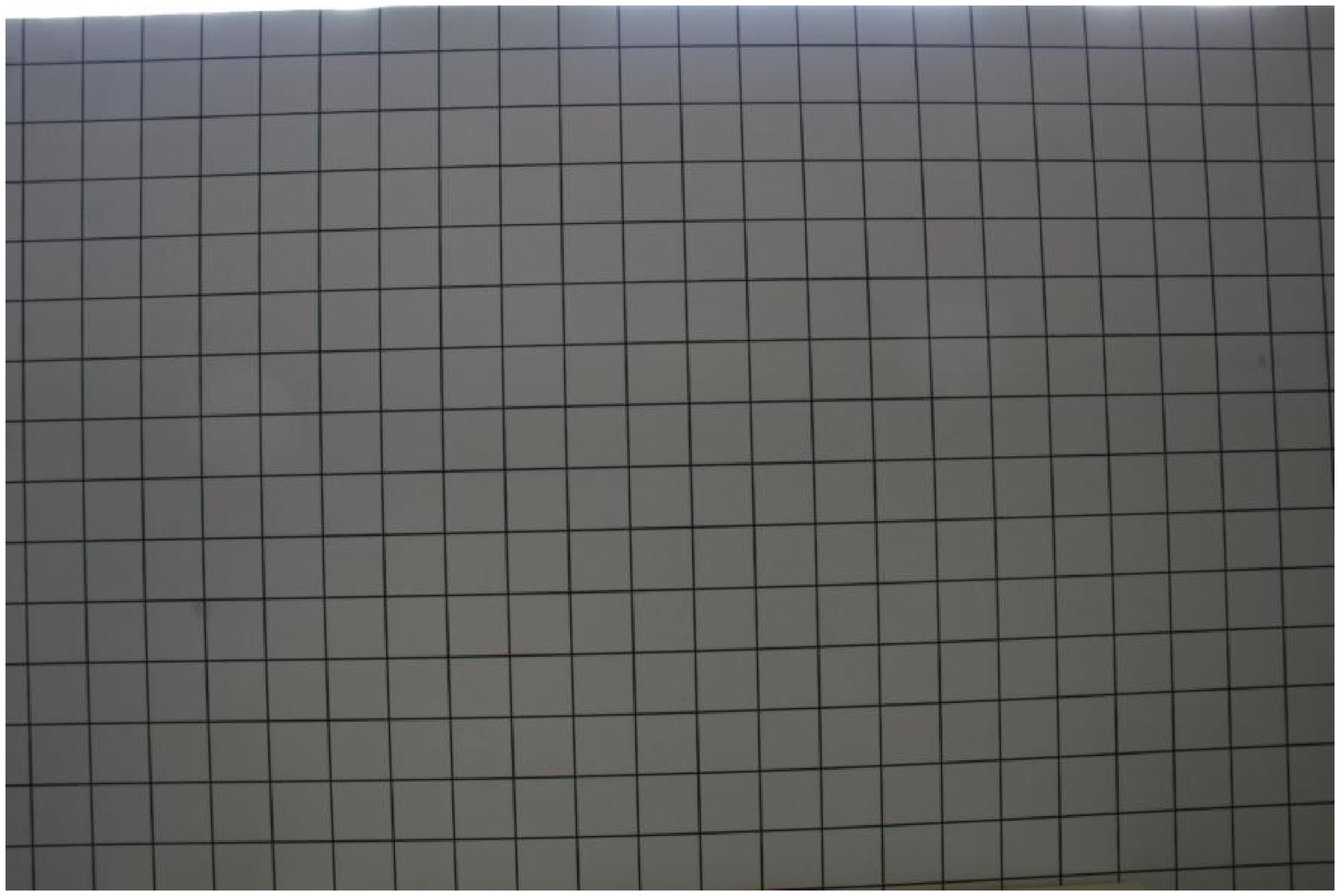}}
\subfloat[Harp image]{\label{subfig:harp_img}
  \includegraphics[width=0.31\textwidth]{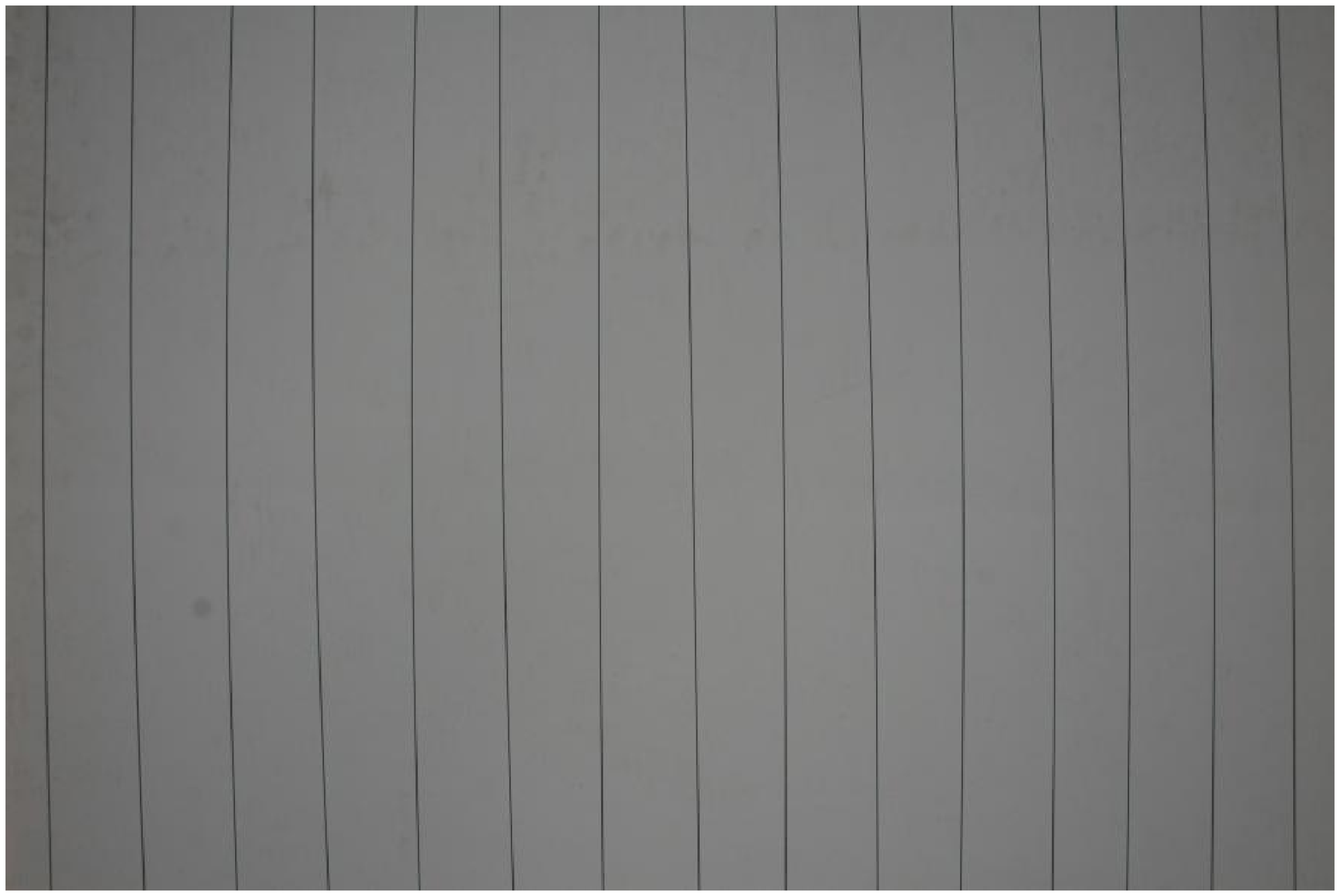}}
\caption{The images used in the Lavest et~al. method.}
\label{fig:three_images_clicks}
\end{figure}

In general, minimizing $E(\mathbf{k})$ is equivalent to solve a set of
equations:
\begin{equation}
 \frac{\partial E(\mathbf{k})}{\partial k_i} = 0, i = 1, 2, \ldots, N.
\end{equation}
When there is only one unknown parameter, the solution can be
approximated by solving the root of the resulting univariate
polynomial. When there are two unknown parameters, resultant-based
method can be used to minimize $E(\mathbf{k})$. The case of more than
two variables requires Gr\"{o}bner basis techniques or
multivariate-resultants based method (see \cite{Alvarez2009} for more
detail). To make the algorithm efficient, \cite{Alvarez2009} optimizes
on two parameters at one time and iterates:
\begin{enumerate}
 \item Obtain distorted edge points which are the 2D projection of 3D
   straight segments;
 \item Initialize $\mathbf{k} = (1,0,...,0)^T$;
 \item Choose any pair $(p , q), 1 \leq p, q \leq N $ and fix all the
   other parameters, then optimize $k_p$ and $k_q$ by resultant-based
   method;
 \item Update $k_0$ using a zoom factor such that distorted and
   undistorted points are as close as possible;
 \item Repeat Step 3 and 4 until all the parameters are estimated.
\end{enumerate}
It is usually supposed that the edge points are already available for
the plumb-line methods. But in practice, it is not a trivial problem
to precisely extract aligned edge points in images. For example, the
online demo \cite{AlvarezIPOL} of Alvarez et~al. method
\cite{Alvarez2009} requires the user to click manually edge
points. This is on the one hand a tedious and time consuming work, and
on the other hand, it may reduce the precision of edge points. Our
method thus gives the possibility to automatize plumb-line methods. We
fed four kinds of edge points to the Alvarez et~al. method: first the
manually clicked edge points of a natural image
(Fig.~\ref{subfig:natural_img}), second the manually clicked edge
points of an image of the grid pattern (Fig.~\ref{subfig:grid_img}),
third the manually clicked edge points of an image of the calibration
harp (Fig.~\ref{subfig:harp_img}) and finally the automatically
extracted edge points of an image of the calibration harp
(Fig.~\ref{subfig:harp_img}), as described in
section~\ref{sec:edge_detection}. These points were used as the input
to the Alvarez et~al. method to estimate an order-4 radial distortion
model, which will be used to correct the distorted images in
Fig.~\ref{fig:distorted_calib_harp_photos}. The precision of this
correction was finally evaluated by the method proposed in the paper
(applied to a different set of images of the calibration harp from the
one used to estimate the correction).

The results in Table~\ref{tab:manual_alvarez_vs_automatic_alvarez}
show that the edge point extraction by our proposed method strengthens
the plumb-line method in terms of precision and spares the long,
tedious and imprecise manual point clicking task. Compared to the
manual clicks with the calibration harp, the improvement in precision
is moderate. Indeed, the Alvarez et~al. method is applied on a very
good quality photograph of the harp. The manual clicks were carefully
placed on the lines across the domain of the image. The slight
inaccuracy of the clicks was smoothed out by our method which applies
a Gaussian convolution of the edge points along the edges, see
section~\ref{sec:edge_detection}. The manual clicks on the image of
the grid pattern and on the natural image give a precision that is two
or three times lower than the calibration harp. For the grid pattern,
the imprecision may come from the non-flatness error, the engraved
straight lines on the pattern being not really straight. For the
natural image, the imprecision comes from two aspects: one is again
the non-straightness error of the lines, the other is the lack of
lines at the border of the image domain, which can explain a precision
decay near the image border.

\begin{table}[!htb]
\caption{The distortion correction performance of the Alvarez
  et~al. method on four kinds of input edge points: manual clicks on
  natural image, manual clicks on a grid pattern image, manual clicks
  on a calibration harp image and automatic edge points extraction on
  the calibration harp image. Compare $d$, $d_{\text{max}}$ and the
  time to obtain the edge points.}
\label{tab:manual_alvarez_vs_automatic_alvarez}
\begin{center}
\begin{tabular}{lccccccc}
\hline
method & time (mins) & $d$ (pixels) & $\bar{d}_{\text{max}}$ \\ 
\hline
Natural image    (manually)      & $\sim$\ 5   & $0.27$ & $1.02$  \\ 
\hline
Grid pattern     (manually)      & $\sim$\ 25  & $0.30$ & $0.94$  \\ 
\hline
Calibration harp (manually)      & $\sim$\ 30  & $0.11$ & $0.39$  \\ 
\hline
Calibration harp (automatically) & $\sim$\ 0   & $0.08$ & $0.27$  \\ 
\hline
\end{tabular}
\end{center}
\end{table}

\section{Conclusion}\label{sec:conclusion}

A ``calibration harp'' has been proposed for camera distortion
measurement, along with its associated image processing chain. This
harp is both a measurement tool and a correction tool. As a
measurement tool, it can be used independently to measure the residual
distortion left by any distortion correction methods or any
software. As a correction tool, the precise edge points detected on
the harp can be used as the input to plumb-line methods and lead to
more precise distortion correction result. In the future, we aim at
finding a more general distortion model to correct more severe
distortion by using the calibration harp. The ideal case would be a
harp-free distortion correction method. But we think that the harp
will always remain useful, as a final measurement tool to validate any
other correction precision and or to detect its failures.

\section*{Acknowledgment}
This work was supported by Agence Nationale de la Recherche
ANR-09-CORD-003 (Callisto project), the MISS project of Centre
National d'Etudes Spatiales, the Office of Naval research under grant
N00014-97-1-0839 and the European Research Council, advanced grant
``Twelve labours''.

\bibliographystyle{osajnl}

\begin{thebibliography}{10}
\newcommand{\enquote}[1]{``#1''}

\bibitem{Slama1980}
C.~Slama, \emph{Manual of Photogrammetry, 4th edition} (Falls Church, American
  Society of Photogrammetry, Virginia, 1980).

\bibitem{Tsai1987}
R.~Tsai, \enquote{{A versatile camera calibration technique for high-accuracy
  3D machine vision metrology using off-the-shelf TV cameras and lenses},} IEEE
  Journal of Robotics and Automation \textbf{3}, 323--344 (1987).

\bibitem{Zhang2000}
Z.~Zhang, \enquote{A flexible new technique for camera calibration,} IEEE
  Transactions on Pattern Analysis and Machine Intelligence pp. 1330--1334
  (2000).

\bibitem{Lavest1998}
J.~Lavest, M.~Viala, and M.~Dhome, \enquote{Do we really need accurate
  calibration pattern to achieve a reliable camera calibration?} ECCV
  \textbf{1}, 158--174 (1998).

\bibitem{Weng1992}
M.~H. J.~Weng, P.~Cohen, \enquote{Camera calibration with distortion models and
  accuracy evaluation,} IEEE Transactions on Pattern Analysis and Machine
  Intelligence \textbf{14}, 965--980 (1992).

\bibitem{TheseZhongwei}
Z.~Tang, \enquote{Calibration de cam{\'e}ra {\`a} haute pr{\'e}cision,} Ph.D.
  thesis, ENS Cachan (2011).

\bibitem{Grompone2010}
R.~Grompone~von Gioi, P.~Monasse, J.-M. Morel, and Z.~Tang, \enquote{Towards
  high-precision lens distortion correction,} ICIP pp. 4237--4240 (2010).

\bibitem{Stein1997}
G.~P. Stein, \enquote{Lens distortion calibration using point correspondences,}
  CVPR pp. 602--608 (1997).

\bibitem{Zhang1996}
Z.~Zhang, \enquote{On the epipolar geometry between two images with lens
  distortion,} ICPR pp. 407--411 (1996).

\bibitem{Fitzgibbon2001}
A.~Fitzgibbon, \enquote{Simultaneous linear estimation of multiple view
  geometry and lens distortion,} CVPR \textbf{1}, 125--132 (2001).

\bibitem{Micusik2003}
B.~Micusik and T.~Pajdla, \enquote{Estimation of omnidirectional camera model
  from epipolar geometry,} CVPR pp. 485--490 (2003).

\bibitem{Li2005}
H.~Li and R.~Hartley, \enquote{A non-iterative method for correcting lens
  distortion from nine-point correspondences,} in \enquote{OmniVision,}
  (2005).

\bibitem{Thirthala2005}
S.~Thirthala and M.~Pollefeys, \enquote{The radial trifocal tensor: a tool for
  calibrating the radial distortion of wide-angle cameras,} CVPR \textbf{1},
  321--328 (2005).

\bibitem{Claus2005a}
D.~Claus and A.~Fitzgibbon, \enquote{A rational function lens distortion model
  for general cameras,} CVPR \textbf{1}, 213--219 (2005).

\bibitem{Barreto2005}
J.~Barreto and K.~Daniilidis, \enquote{Fundamental matrix for cameras with
  radial distortion,} ICCV \textbf{1}, 625--632 (2005).

\bibitem{Kukelova2007}
Z.~Kukelova and T.~Pajdla, \enquote{Two minimal problems for cameras with
  radial distortion,} ICCV pp. 1--8 (2007).

\bibitem{Kukelova2008}
Z.~Kukelova, M.~Bujnak, and T.~Pajdla, \enquote{Automatic generator of minimal
  problem solvers,} ECCV pp. 302--315 (2008).

\bibitem{Byrod2008}
M.~Byrod, Z.~Kukelova, K.~Josephson, T.~Pajdla, and K.~Astrom, \enquote{Fast
  and robust numerical solutions to minimal problems for cameras with radial
  distortion,} CVPR pp. 1--8 (2008).

\bibitem{Kukelova2007a}
Z.~Kukelova and T.~Pajdla, \enquote{A minimal solution to the autocalibration
  of radial distortion,} CVPR pp. 1--7 (2007).

\bibitem{Josephson2009}
K.~Josephson and M.~Byrod, \enquote{Pose estimation with radial distortion and
  unknown focal length,} CVPR pp. 2419--2426 (2009).

\bibitem{Triggs2000}
B.~Triggs, P.~Mclauchlan, R.~Hartley, and A.~Fitzgibbon, \enquote{Bundle
  adjustment -- a modern synthesis,} Vision Algorithms: Theory and Practice,
  LNCS pp. 298--375 (2000).

\bibitem{Brown1971}
D.~Brown, \enquote{Close-range camera calibration,} Photogrammetric Engieering
  \textbf{37}, 855--866 (1971).

\bibitem{Alvarez2009}
L.~Alvarez, L.~Gomez, and J.~Rafael~Sendra, \enquote{An algebraic approach to
  lens distortion by line rectification,} Journal of Mathematical Imaging and
  Vision \textbf{35(1)}, 36--50 (2009).

\bibitem{PRESCOTT1997}
B.~Prescott and G.~McLean, \enquote{Line-based correction of radial lens
  distortion,} Graphical Models and Image Processing \textbf{59}, 39--47
  (1997).

\bibitem{Pajdla1997}
T.~Pajdla, T.~Werner, and V.~Hlavac, \enquote{Correcting radial lens distortion
  without knowledge of {3-D} structure,} Research Report, Czech Technical
  University  (1997).

\bibitem{Devernay2001}
F.~Devernay and O.~Faugeras, \enquote{Straight lines have to be straight,}
  Machine Vision and Applications \textbf{13}, 14--24 (2001).

\bibitem{Claus2005}
D.~Claus and A.~Fitzgibbon, \enquote{A plumbline constraint for the rational
  function lens distortion model,} BMVC pp. 99--108 (2005).

\bibitem{Hartley2004}
R.~Hartley and A.~Zisserman, \emph{Multiple View Geometry in Computer Vision}
  (Cambridge University Press, 2004).

\bibitem{Rosten2011}
E.~Rosten and R.~Loveland, \enquote{Camera distortion self-calibration using
  the plumb-line constraint and minimal hough entropy,} Machine Vision and
  Applications \textbf{22}, 77--85 (2011).

\bibitem{Devernay1995}
F.~Devernay, \enquote{A non-maxima suppression method for edge detection with
  sub-pixel accuracy,} Tech. Rep. 2724, INRIA rapport de recherche (1995).

\bibitem{Canny1986}
J.~Canny, \enquote{A computational approach to edge detection,} IEEE
  Transactions on Pattern Analysis and Machine Intelligence \textbf{8},
  679--698 (1986).

\bibitem{Deriche1987}
R.~Deriche, \enquote{Using {C}anny's criteria to derive a recursively
  implemented optimal edge detector,} International Journal of Computer Vision
  \textbf{1}, 167--187 (1987).

\bibitem{Grompone2010lsd}
R.~Grompone~von Gioi, J.~Jakubowicz, J.~Morel, and G.~Randall, \enquote{{LSD}:
  A fast {L}ine {S}egment {D}etector with a false detection control,} IEEE
  Transactions on Pattern Analysis and Machine Intelligence \textbf{32},
  722--732 (2010).

\bibitem{IPOLLSDpaper}
R.~Grompone~von Gioi, J.~Jakubowicz, J.~Morel, and G.~Randall, \enquote{{LSD: a
  Line Segment Detector},} Image Processing On Line  (2012).
  DOI:\url{http://dx.doi.org/10.5201/ipol.2012.gjmr-lsd}.

\bibitem{Morel2011sift}
J.~Morel and G.~Yu, \enquote{{Is SIFT scale invariant?}} Inverse Problems and
  Imaging \textbf{5}, 115--136 (2011).

\bibitem{AlvarezIPOL}
L.~G. Luis~Alvarez and J.~R. Sendra, \enquote{{Algebraic Lens Distortion Model
  Estimation},} Image Processing On Line  (2010).

\end{thebibliography}

\end{document}